%% file: main.tex
\documentclass[10pt,twocolumn,letterpaper]{article}

\usepackage{cvpr}              
\input{preamble}
\definecolor{cvprblue}{rgb}{0.21,0.49,0.74}
\usepackage[pagebackref,breaklinks,colorlinks,allcolors=cvprblue]{hyperref}

\usepackage{makecell}
\usepackage{multirow}
\usepackage{colortbl}
\definecolor{mypink2}{rgb}{.99,.96,.98}
\definecolor{mypink1}{rgb}{.99,.93,.98}
\definecolor{mypink}{rgb}{.99,.90,.98}
\usepackage{algorithm}
\usepackage{algpseudocode}
\usepackage{multirow}
\usepackage{indentfirst}
\usepackage{float}
\usepackage{fontawesome}

\def\paperID{18386} 
\def\confName{CVPR}
\def\confYear{2026}

\title{Mind the Generative Details: Direct Localized Detail Preference Optimization for Video Diffusion Models}

\author{Zitong Huang$^{1,\ast}$ \quad Kaidong Zhang$^{2,\ast}$ \quad Yukang Ding$^{2,\ast,\dagger}$ \quad Chao Gao$^2$\quad Rui Ding$^2$\\
Ying Chen$^2$\quad Wangmeng Zuo$^{1,\dagger}$\\
\\
$^1$Harbin Institute of Technology \\ $^2$Alibaba Group - Taobao \& Tmall Group\\
{\tt\small zitonghuang99@gmail.com}
}

\begin{document}
\maketitle

\renewcommand{\thefootnote}{\fnsymbol{footnote}}

\setcounter{footnote}{1} 
\footnotetext{Equal contribution}

\setcounter{footnote}{2}
\footnotetext{Corresponding author}

\input{sec/0_abstract}

\input{sec/1_intro}

\input{sec/2_related_work}

\input{sec/3_Preliminaries}
\input{sec/4_Proposed_Method}
\input{sec/5_Experiments}

\input{sec/6_Conclusion}

\input{sec/X_suppl}

{
    \small
    \bibliographystyle{ieeenat_fullname}
    \bibliography{main}
}

\end{document}

%% file: sec/0_abstract.tex
\begin{abstract}
Aligning text-to-video diffusion models with human preferences is crucial for generating high-quality videos. Existing Direct Preference Otimization (DPO) methods rely on multi-sample ranking and task-specific critic models, which is inefficient and often yields ambiguous global supervision. To address these limitations, we propose \textbf{LocalDPO}, a novel post-training framework that constructs localized preference pairs from real videos and optimizes alignment at the spatio-temporal region level. We design an automated pipeline to efficiently collect preference pair data that generates preference pairs with \textbf{a single inference} per prompt, eliminating the need for external critic models or manual annotation. Specifically, we treat high-quality real videos as positive samples and generate corresponding negatives by locally corrupting them with random spatio-temporal masks and restoring only the masked regions using the frozen base model. During training, we introduce a region-aware DPO loss that restricts preference learning to corrupted areas for rapid convergence.
Experiments on Wan2.1 and CogVideoX demonstrate that LocalDPO consistently improves video fidelity, temporal coherence and human preference scores over other post-training approaches, establishing a more efficient and fine-grained paradigm for video generator alignment. The code is available at \url{https://github.com/1170300714/Local-DPO}.
\end{abstract}

%% file: sec/1_intro.tex
\vspace{-0.5cm}
\section{Introduction}
\label{sec:intro}

\begin{figure}[!htbp]
    \centering
    \vspace{-0.9cm}
    \includegraphics[width=0.5\textwidth]{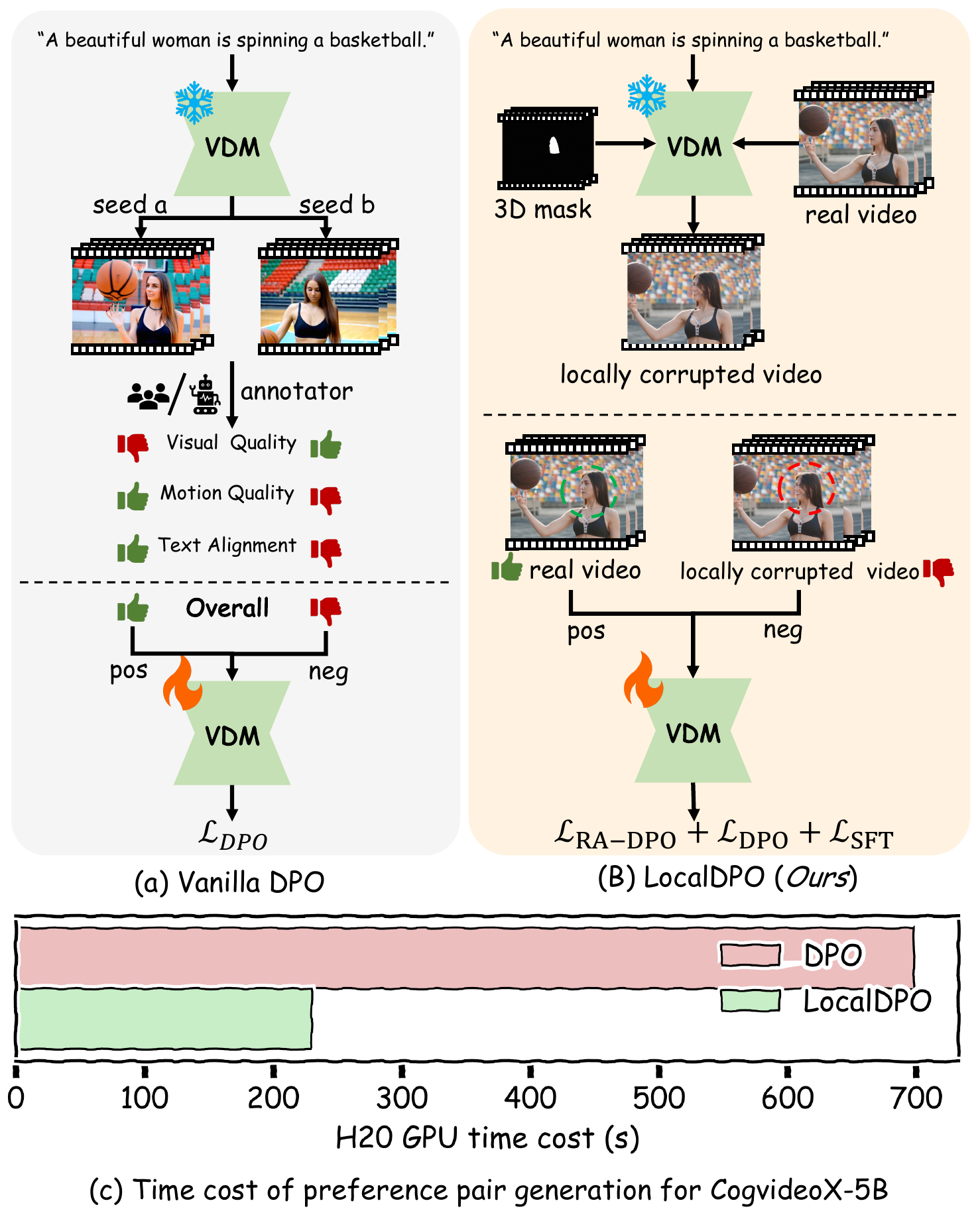} 
    \vspace{-0.9cm}
    \caption{Comparison between (a) vanilla DPO and (b) LocalDPO for video diffusion model (VDM). LocalDPO efficiently constructs positive-negative pairs by locally corrupting real videos, avoiding multi-round sampling, extra critic models, and annotation ambiguities.
    (c) Quantifies comprison of GPU time in constructing preference pairs.
    }
    \vspace{-0.5cm}
    \label{fig:teaser}
\end{figure}

\begin{figure*}[!htbp]
    \centering
    \vspace{-0.8cm}
    \includegraphics[width=1.0\textwidth]{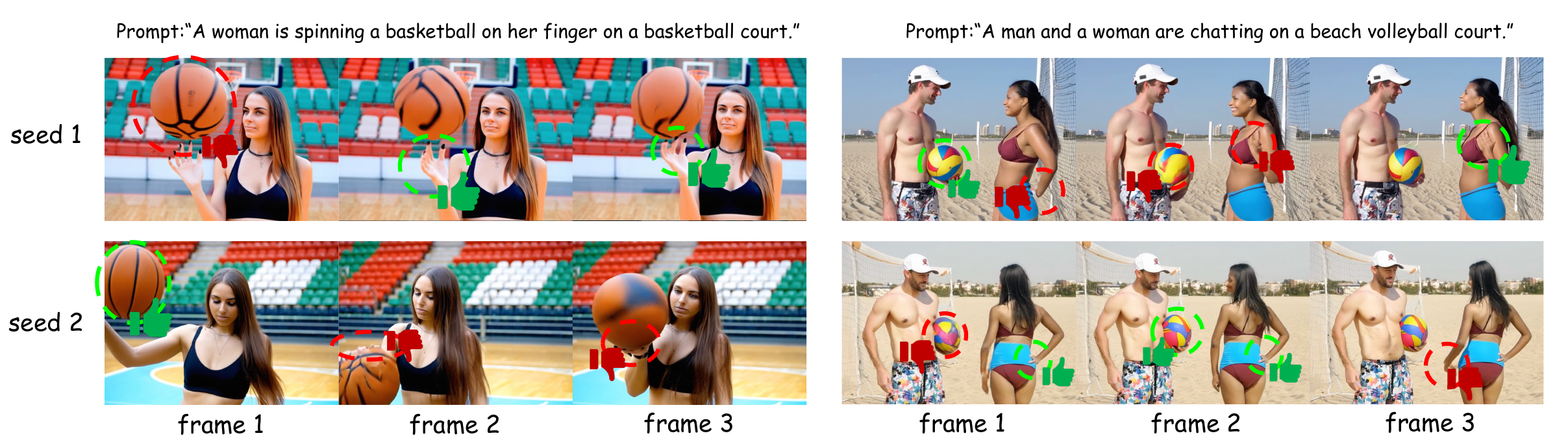} 
    \vspace{-0.8cm}
    \caption{Comparison of video pairs generated by CogVideoX-5B from the same prompt but different seeds reveals significant discrepancies in the visual quality of localized regions, with their relative quality varying across frames. These fine-grained, localized preference patterns are overlooked by the vanilla DPO annotation paradigm, motivating our LocalDPO approach.
    }
    \vspace{-0.5cm}
    \label{fig:prefer}
\end{figure*}

Recent advances in diffusion models~\citep{croitoru2023diffusion,yang2023diffusion,ho2020denoising, ddpm, ddim, flowmatching} have enabled impressive progress in text-to-video generation, where the goal is to synthesize temporally coherent and semantically aligned videos from language prompts. 
Despite the success of large-scale pre-trained video diffusion models (VDMs)~\citep{ho2022video,blattmann2023stable, xing2024survey, wan2.1, hunyuanvideo, seawead2025seaweed, yang2024cogvideox, hong2022cogvideo}, generated videos often suffer from artifacts such as flickering objects, inconsistent motions, or implausible local details. 
A straightforward approach to further improve generation quality is supervised fine-tuning (SFT) on curated collections of high-quality real videos, which directly aligns the model with human-preferred outputs. 
However, SFT treats all training samples equally and lacks an explicit mechanism to learn from relative quality differences, making it insensitive to subtle but perceptually critical artifacts, such as flickering objects or inconsistent motions.
To address this limitation, recent work has turned to preference-based alignment~\citep{rafailov2023direct,schulman2017proximal}, particularly Direct Preference Optimization (DPO)~\citep{rafailov2023direct}, which fine-tunes the model using annotated pairwise preference data. 
This training paradigm enables the model to further align with human preferences while also perceiving and avoiding undesirable distributions, which has become a popular and widely adopted post-training technique for video diffusion models.

However, existing video DPO approaches~\citep{videodpo, videoalign, lift, densedpo} still present several crucial limitations that remain to be addressed.
(1) They require generating multiple videos per prompt and ranking them using human annotations or a fine-tuned critic model ~\cite{qwen2.5-vl, videoscore, visionreward, videoalign}. 
This leads to heavy model-inference and high annotation cost.
(2) Preference pairs are typically based on overall scores that aggregate multiple quality dimensions.
However, a video with a higher total score may perform poorly in specific aspects (see Fig.~\ref{fig:teaser} ). 
This can yield ambiguous or even conflicting supervision signals during fine-tuning, thereby impeding model convergence.
(3) Scoring is performed at the global video level,
ignoring region-specific preference cues (such as localized artifacts and detail richness of objects, see Fig.~\ref{fig:prefer}), which are critical to human subjective perception.

To overcome these limitations, we propose LocalDPO, an efficient preference optimization approach that achieves preference learning at the level of local video details, as shown in Fig.~\ref{fig:teaser} (b).
Instead of generating multiple videos and relying on human or model-based annotations, LocalDPO directly uses high-quality real videos as positive samples and corrupts local regions of these videos using the model to be optimized, thereby generating corresponding negative samples with only single inference per prompt.
Specifically, we first propose a random spatio-temporal mask generation algorithm to select the regions to be corrupted. 
This algorithm constructs closed regions by randomly generating multiple Bézier curves in the video, with each curve connected end-to-end to form a loop.
Next, we propose a spatio-temporal local corruption method based on the pre-trained (to-be-optimized) VDM to achieve localized corruption.
This method redraws video content by first adding noise to the original video and then denoising it, while using the mask generated in the previous step to restrict the restoration to specific regions, thereby producing a negative sample that preserves global semantics but exhibits localized degradation.
Finally, we extend the vanilla diffusion DPO loss to a mask-guided regoin-aware DPO loss, which explicitly encourages the model to perform preference optimization in the local regions of the positive–negative sample pairs. This region-aware DPO loss formulation effectively accelerates model convergence.

Our LocalDPO effectively addresses the aforementioned limitations of existing DPO methods:  
(1) \textbf{Low Cost and High Confidence}: LocalDPO uses real videos as positive samples and their corrupted versions as negative samples.
This construction of preference pairs is highly direct and eliminates the need—present in conventional DPO—to first generate multiple videos and then annotate them, thereby saving substantial labeling costs.
Fig.~\ref{fig:teaser} (c) illustrates that LocalDPO clearly outperforms DPO in terms of time cost for constructing preference data.
Negative samples in LocalDPO are produced by the model’s own restoration process, and their quality is inherently lower than that of high-quality real videos in all dimensions. 
Thus, \textit{the resulting preference pairs exhibit consistent superiority of the positive sample over the negative one in every quality aspect}.
(2) \textbf{Localized Fine-Grained Preference Optimization}: The locally corrupted regions and their original counterparts in the real video naturally form fine-grained, region-level preference pairs, enabling the model to explicitly enhance its capacity for local-region preference optimization. These locally degraded negatives exhibit spatial detail loss or collapse and temporal flicker and incoherence, enabling our preference learning to concentrate on generative details.

Quantitative evaluations demonstrate that LocalDPO outperforms SFT, Vanilla DPO and other post-training approaches, producing videos with higher visual fidelity and stronger semantic alignment with the input prompts. 
Furthermore, qualitative assessments reveal that videos generated by LocalDPO exhibit richer, more realistic local details, underscoring the effectiveness of our localized preference optimization strategy.

In a nutshell, the main contributions of this paper are summarized as follows:
\begin{itemize}
    \item We propose LocalDPO, a novel preference optimization method that builds training pairs from real videos and their locally corrupted versions, bypassing costly multi-sample generation and annotations in existing methods. The negative samples are homologous with model and each perference pair is high-confidence.
    \item We propose a mask-guided local region-aware DPO loss to enable fine-grained preference learning on region-level degradations while preserving global coherence.
    \item Extensive experiments show that LocalDPO outperforms pre-trained VDMs, SFT, and existing preference-based methods, producing videos with higher visual fidelity, fewer temporal artifacts, and stronger alignment with input prompts quantitatively and qualitatively.
\end{itemize}


%% file: sec/2_related_work.tex
\vspace{-0.2cm}
\section{Related Work}
\label{sec:formatting}

\subsection{Video Diffusion Model}
Diffusion-based models~\cite{ddpm, ddim, flowmatching, song2020score} have emerged as the dominant paradigm for text-to-video generation, building upon advances in image synthesis~\cite{sdxl, stablediffusion, dall-e2, imagen}. Early approaches extended image diffusion frameworks to the temporal domain via 3D convolutions or recurrent structures, enabling basic text-conditioned video synthesis with coherent motion~\cite{vdm, wu2023tune, khachatryan2023text2video}. Subsequent works improved fidelity, duration, and efficiency through architectural innovations: spatial-temporal U-Nets~\cite{unet, guo2023animatediff, blattmann2023align}, cascaded super-resolution pipelines~\cite{realesrgan, ho2022imagenvideo, wang2025lavie}, and latent-space factorization~\cite{zhou2022magicvideo, wan2.1, hunyuanvideo}. Recently, DiT-based architectures~\cite{dit, vit, sd3} combined with 3D-VAEs~\cite{yang2024cogvideox, hunyuanvideo, wan2.1} have become prevalent, leveraging cross-modal attention~\cite{attention} to enhance temporal coherence, motion plausibility, and semantic alignment.

Despite these advances, current methods still suffer from generation failures such as temporal flickering, implausible motion, visual artifacts, or poor text alignment~\cite{ling2025vmbench, vbench, chefer2025videojam}. A common remedy is supervised fine-tuning on large-scale, high-quality datasets~\cite{hu2022lora}, yet this demands extensive data collection~\cite{chen2024panda, bain2021frozen, nan2024openvid} and annotation~\cite{chen2024internvl, wang2024cogvlm}, while still struggling with persistent issues like scene transitions and watermarks~\cite{hong2022cogvideo, yang2024cogvideox}.

\subsection{Preference Learning for Video Generation}
Direct Preference Optimization (DPO)~\cite{rafailov2023direct}, a prominent alignment technique from large language models, offers a training strategy based solely on curated positive–negative pairs, avoiding explicit reward modeling and mitigating issues like reward hacking in RLHF~\cite{schulman2017proximal, zheng2023secrets}. Since~\cite{diffusionDPO} first extended DPO to diffusion models for text-to-image synthesis, preference optimization has been increasingly adopted for video generation~\cite{lift, densedpo, videodpo, personalvideo, videoalign, yang2025ipo}. For instance,~\cite{lift} trains a reward model on human-curated data to refine T2V models via reward-weighted likelihood;~\cite{videodpo} constructs preference scores for pairwise data collection to improve visual quality and alignment;~\cite{videoalign} leverages multi-dimensional evaluation and flow-based alignment to enhance generation capability.

However, existing video DPO methods predominantly rely on multi-sample ranking, where differences between videos are often global, inconsistent, or dominated by stochastic noise rather than interpretable quality degradation. More critically, these approaches overlook local failure modes—such as flickering objects or distorted regions—that disproportionately affect human perception. This limitation weakens the learning signal and hinders fine-grained quality control. Our work addresses these issues by constructing preference pairs with controlled, localized corruptions and explicitly optimizing alignment within the affected spatio-temporal regions.

%% file: sec/3_Preliminaries.tex
\begin{figure*}[!htbp]
    \centering
    \vspace{-0.9cm}
    \includegraphics[width=1.0\textwidth]{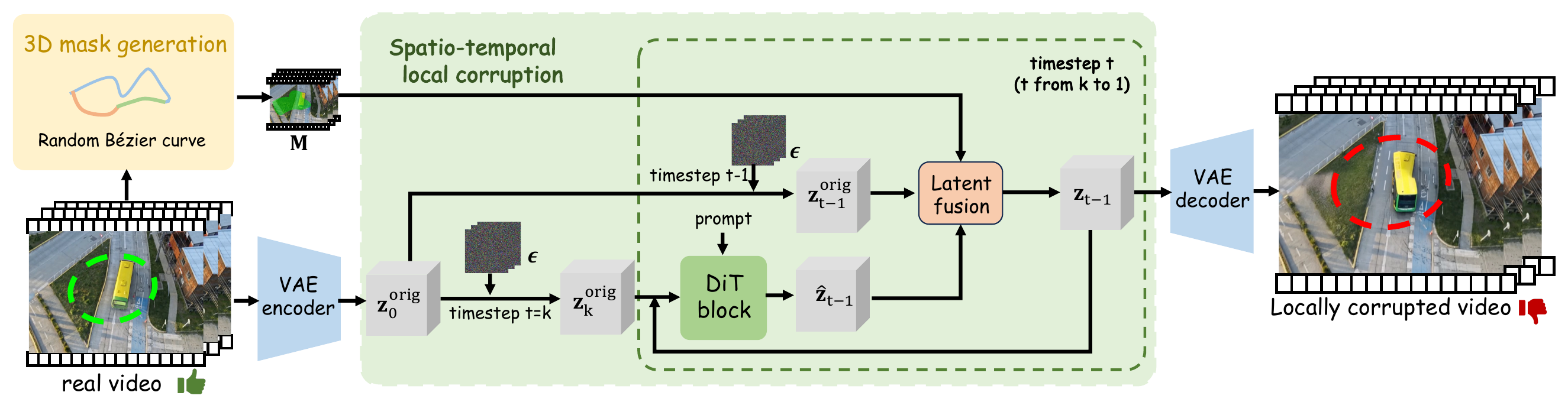} 
    \vspace{-0.7cm}
    \caption{\textbf{Pipeline of locally corrupted videos generation.}
     We first randomly sample several Bézier curves on the original video and ensure that these curves form closed shapes. 
    The interior of each closed shape defines the region to be corrupted in subsequent steps.
Then, the masked area of real video is inpainted by the pretrained VDM.
Specifically, given the latent of input real video, the model first adds a controlled amount of noise to its latent representation and then  denoises it step by step.
During each denoising step, the original video latent is re-noised at the noise level corresponding to the next timestep and then fused with the denoised latent via a \textbf{latent fusion mechanism} by $\mathbf{z}_{t-1} = \mathbf{M} \odot \hat{\mathbf{z}}_{t-1} + (1 - \mathbf{M}) \odot \mathbf{z}_{t-1}^{\text{orig}}$.
    }
    \vspace{-0.5cm}
    \label{fig:pipeline}
\end{figure*}

\vspace{-0.2cm}
\section{Preliminaries}
\label{sec:preliminaries}

\noindent \textbf{Diffusion DPO for Video Generation Models}. 
Direct Preference Optimization (DPO) has been extended to latent diffusion models for video generation by operating entirely in the latent space, where it aligns the generative model with human preferences by encouraging lower prediction errors (e.g., in noise or velocity) on preferred videos compared to dispreferred ones.
Formally, given an annotated preference dataset $\mathcal{D} = \{(\mathbf{c}, \mathbf{x}^w, \mathbf{x}^l)\}$, where $\mathbf{c}$ is a text prompt and $\mathbf{x}^w, \mathbf{x}^l \in \mathbb{R}^{T \times H \times W \times C}$ are the preferred and dispreferred videos, a pretrained 3D variational autoencoder (VAE) encoder ~\cite{vae, yang2024cogvideox, wan2.1} $\text{Enc}(\cdot)$ maps them to latent representations $\mathbf{z}^w = \text{Enc}(\mathbf{x}^w)$ and $\mathbf{z}^l = \text{Enc}(\mathbf{x}^l)$, with $\mathbf{z} \in \mathbb{R}^{T' \times H' \times W' \times C'}$.
Let $f_\theta(\cdot, t, \mathbf{c})$ denote the noise predictor (for DDPM based model) or velocity estimator (for rectified-flow based model) of the diffusion model to be optimized, and $f_{\tilde{\theta}}(\cdot, t, \mathbf{c})$ stands for a corresponding fixed reference model.
For each preference pair $(\mathbf{z}^w, \mathbf{z}^l)$ under prompt $\mathbf{c}$, DPO minimizes the following loss:
{\small
\begin{equation}
    \mathcal{L}_{\text{DPO}} = -\mathbb{E}_{(\mathbf{c}, \mathbf{z}^w, \mathbf{z}^l) \sim \mathcal{D}} \left[ \log \sigma \left( - \beta \cdot \mathbb{E}_{t} \left[ 
        \Delta_w - \Delta_l
    \right] \right) \right],
\end{equation}
}
where $\sigma(\cdot)$ is the sigmoid function, $\beta > 0$ is the temperature, and $\mathbf{y}^{*}$ denotes the corresponding ground-truth target ($\epsilon$ for DDPM based methods or $\epsilon-\mathbf{z}$ for rectified-flow based methods).
$\Delta_w$ and $\Delta_l$ are the abbreviations of 
$\Delta(\mathbf{z}^w, t, \mathbf{c},\mathbf{y}^w)$ and $\Delta(\mathbf{z}^l, t, \mathbf{c},\mathbf{y}^l)$, 
where $\Delta_*$  measures the improvement of the current model over the reference model in terms of reconstruction error on latent $\mathbf{z}$ at timestep $t$:
{\small
\begin{align}
    \Delta(\mathbf{z}^*, t, \mathbf{c},\mathbf{y}^*) = 
    &\left\| \mathbf{y}^* - f_{\theta}(\mathbf{z}_t^*, t, \mathbf{c}) \right\|^2 - \left\| \mathbf{y}^* - f_{\tilde{\theta}}(\mathbf{z}_t^*, t, \mathbf{c}) \right\|^2,
\end{align}
}
with $\mathbf{z}_t$ denoting the noisy version of $\mathbf{z}$ at timestep $t$, and $\mathbf{y}$ representing the ground-truth noise used to construct $\mathbf{z}_t$.  

\noindent \textbf{Limitation of Diffusion DPO.}
Despite its elegance and empirical effectiveness, current video DPO approaches~\citep{videodpo,videoalign} suffer from several practical and conceptual limitations that hinder their scalability and alignment fidelity.
First, they typically require generating multiple candidate videos per prompt and obtaining human or reward-model-based rankings—a process that incurs high annotation costs. 
Second, preferences are usually derived from global quality scores that aggregate diverse aspects (e.g., motion smoothness, visual fidelity, semantic alignment).
However, a video with a higher aggregate score may underperform in specific perceptually critical dimensions, leading to ambiguous or even conflicting supervision signals during fine-tuning. 
Third, existing methods treat videos as monolithic entities and ignore localized preference cues—such as facial artifacts or object distortions. \textit{These shortcomings motivate the development of a more efficient DPO framework—one that constructs preference pairs more effectively, enforces stronger preference consistency, and explicitly accounts for region-level perceptual preferences.}

%% file: sec/4_Proposed_Method.tex
\vspace{-0.2cm}
\section{Methodology}
\label{sec:method}

\subsection{Overview}

This paper proposes LocalDPO, which addresses the aforementioned limitations of existing DPO methods through the following key ideas.
To improve the efficiency of preference pair construction, LocalDPO innovatively uses high-quality real videos as preferred samples and generates dispreferred samples by applying localized corruption to these real videos. 
This strategy drastically reduces the number of videos that need to be generated and eliminates the need for human or reward-model-based labeling, enabling highly efficient preference pair creation.
To ensure preference consistency, LocalDPO leverages the fact that videos with localized corruptions are inherently of lower quality than their original high-quality counterparts, guaranteeing a reliable and unambiguous preference order within each pair.
Finally, to better capture region-level perceptual preferences, LocalDPO introduces a region aware DPO loss that explicitly encourages the model to refine fine-grained details in specific spatial regions. 
The whole pipeline of LocalDPO is shown in Fig.~\ref{fig:pipeline}.
In Sec.~\ref{subsec:video_generation}, we will detail how a pretrained video diffusion model (VDM) is employed to corrupted local regions of real videos, thereby generating dispreferred samples. 
Sec.~\ref{subsec:local_dpo} will describe how the resulting preference dataset is utilized to enhance the model’s ability to align with human preferences at the level of local visual details.

\vspace{-0.2cm}
\subsection{Locally Corrupted Videos Generation}
\label{subsec:video_generation}
An illustration of locally corrupted videos generation is shown in Fig.~\ref{fig:pipeline}.
Given a real video $\mathbf{x}^w$ and its corresponding text prompt $\mathbf{c}$, our goal is to generate a dispreferred video $\mathbf{x}^l$ such that $\mathbf{x}^l$ is the degradation w.r.t. $\mathbf{x}^w$ only in a localized region, thereby forming a region-aware preference tuple $(\mathbf{c}, \mathbf{x}^w, \mathbf{x}^l, \mathbf{M})$, where $\mathbf{M} \in \{0, 1\}^{T' \times H' \times W'}$ denotes the binary mask indicating the corrupted regions.
To achieve this goal, two sub-problems are necessary to be addressed: (1) how to select the regions to be corrupted (i.e., how to obtain \(\mathbf{M}\)), and (2) how to generate corruption within those regions that reflects the inherent generative bias of the policy model.

\noindent\textbf{3D Mask Generation.} 

This paper adopts a simple yet efficient strategy to select regions for corruption: we randomly generate irregular closed shapes in the spatial domain of the video. 
We propose a randomized closed-shape generation algorithm based on Bézier curves. 
Specifically, we sequentially generate $P$ Bézier curves within the spatial extent of the current video.
Initially, a set of control points is generated within the first video frame. These points are subsequently connected using cubic Bézier curves to form a closed, cyclic contour.
We then impose random rotation and movement to broadcast the initial Bézier curves across all the subsequent frames at the corresponding spatial location to construct a 3D spatio-temporal mask, which is subsequently downsampled according to the VAE's downsampling factor to obtain the final \(\mathbf{M}\).
\textit{The formal algorithm is provided in the supplementary material.}

\noindent\textbf{Spatio-temporal Local Corruption}.
To generate a disprefered sample $\mathbf{x}^l$ that degrades only within the masked region $M$ while preserving the original content elsewhere, we perform a masked progressive denoising process using the pretrained VDM.
Let $\mathbf{z}_0^{\text{orig}} = \mathcal{E}(\mathbf{x}_w)$ denote the clean latent of the real video. 
We first sample a noise level $\alpha \in [\alpha_l, \alpha_h]$, where $0 < \alpha_l < \alpha_h < 1$ are two hyperparameters.
We use this noise level to add noise to $z_0$ and obtain $z_k$, where $k = \lceil T \times \alpha \rceil$ denotes the timestep and $T$ is the total number of denoising steps, typically set to $1,000$, We denoise the $z_k$ from $t = 0$ iteratively to obtain the local corruption sample $x_l$.
After each denoising step, a region-aware latent fusion mechanism is performed to ensure that only the latents within the masked region are corrupted, where we retain only the denoised latents inside the mask but the content outside the mask is replaced with the re-noised version of the original video latent at the next timestep, thereby forming the final output of that denoising step.
Formally, given the current noisy latent $\mathbf{z}_t$, the model produces a denoised estimation $\hat{\mathbf{z}}_{t-1} = f_{\theta}(\mathbf{z}_t, t, \mathbf{c})$. 
Then, the original clean latent $\mathbf{z}_0^{\text{orig}}$ is re-noised to timestep $t-1$, and the region-aware latent fusion is devised as:
{\small
\begin{equation}
    \mathbf{z}_{t-1} = \mathbf{M} \odot \hat{\mathbf{z}}_{t-1} + (1 - \mathbf{M}) \odot \mathbf{z}_{t-1}^{\text{orig}}.
\end{equation}
}
Where $\odot$ denotes the Hadamard (element-wise) product.
This procedure guarantees that at each step, the latents in both masked and unmasked regions retains to the same noise level, thereby avoiding distributional mismatch that causes denoising failure.
After completing the denoising trajectory, we obtain the final disprefered latent $\mathbf{x}^l = \text{Dec}(\mathbf{z}_0)$, which is identical to $\mathbf{x}^w$ outside $\mathbf{M}$ but contains model-synthesized and corrupted content inside $\mathbf{M}$.
The resulting pair $(\mathbf{c}, \mathbf{x}^w, \mathbf{x}^l, \mathbf{M},\alpha)$ thus provides a unambiguous, localized preference signal for training.

\subsection{Region Aware Preference Optimization}
\label{subsec:local_dpo}
We expect the model to fully capture the divergence in the corrupted regions between positive and negative samples in the preference dataset $ \mathcal{\hat{D}} = \{(\mathbf{c}, \mathbf{x}^w, \mathbf{x}^l, \mathbf{M},\alpha)_i\}^N_{i=1}$.
Therefore, we design a method to extend the vanilla diffusion DPO loss into a \textit{region-aware preference optimization objective, denoted by $\mathcal{L}_\text{RA-DPO}$}:
{\scriptsize
\begin{equation}
\begin{split}
    \mathcal{L}_{\text{RA-DPO}} = -\mathbb{E}_{d \sim \mathcal{\hat{D}}} \Bigg[ \log \sigma \Bigg( - \beta \cdot (1+\eta(\alpha))  \cdot \mathbb{E}_{t} \big[ 
        \Delta'_w  -  \Delta'_l
    \big] \Bigg) \Bigg],
\end{split}
\end{equation}
} 
where $d\triangleq(\mathbf{c}, \mathbf{x}^w, \mathbf{x}^l, \mathbf{M},\alpha)$ represents the data sample from the preference dataset $\mathcal{\hat{D}}$ and $\eta(\alpha) = \frac{\alpha-\alpha_l}{\alpha_h-\alpha_l}$ is the normalization function used to normalize noise level $\alpha$ for optimization, dynamically adjusting the strength of the penalty based on the degree of corruption. 
$\Delta'_*$ is the abbreviation of $\Delta'(\mathbf{z}^*, t, \mathbf{c},\mathbf{y}^*, \mathbf{M})$ which measures the improvement of the current model over the reference model in terms of reconstruction error in $\mathbf{M}$ on latent $\mathbf{z}$ at timestep $t$:
{\scriptsize
\begin{align}
    \Delta'_* = \frac{N_{M}}{||\mathbf{M}||_1}(
    &\left\| \mathbf{M} \odot(\mathbf{y}^* - f_{\theta}(\mathbf{z}_t^*, t, \mathbf{c})) \right\|^2 - \left\| \mathbf{M} \odot (\mathbf{y}^* - f_{\tilde{\theta}}(\mathbf{z}_t^*, t, \mathbf{c})) \right\|^2),
\end{align}
}
where $N_{M} = T' \times H' \times W'$ indicates the total  number of elements in the $\mathbf{M}$.

\noindent\textbf{Hybrid training objective.} Excessively prioritizing local pairwise preferences may lead to overfitting and impair the model’s overall capacity to capture global video structure.
To address this issue, we incorporate the standard diffusion DPO and supervised fine-tuning (SFT) losses as regularization terms during training, thereby promoting stable and robust optimization.
{\small
\begin{equation}
    \mathcal{L}_{\text{total}} = \lambda_{\text{RA-DPO}} \mathcal{L}_{\text{RA-DPO}} + \lambda_{\text{DPO}} \mathcal{L}_{\text{DPO}} + \lambda_{\text{SFT}} \mathcal{L}_{\text{SFT}},
\end{equation}
}
where $\mathcal{L}_{\text{DPO}}$ is the standard diffusion DPO loss applied to the full latent (\ie, with $\mathbf{M} \equiv 1$); $\mathcal{L}_{\text{SFT}} = \mathbb{E}_{t} \left[ \left\| \mathbf{y}^{w} - f_{\theta}(\mathbf{z}^w_t, t, \mathbf{c}) \right\|^2 \right]$ is the supervised fine-tuning loss on real video latents, which anchors the model to high-quality data; $\lambda_{\text{RA-DPO}}, \lambda_{\text{DPO}}, \lambda_{\text{SFT}}$ are coefficients.
This design enables LocalDPO to learn fine-grained, region-specific alignment while preserving the global capabilities of the base model.

%% file: sec/5_Experiments.tex
\vspace{-0.2cm}
\section{Experiments}

\subsection{Datasets}
Following the data-construction pipeline~\cite{wang2025koala, ju2024miradata, wan2.1} and filtering protocols~\cite{tschannen2025siglip, wei2024got, soucek2024transnet, schuhmann2021laion, raft, ke2021musiq, dover}, we curate a large dataset containing initial video clips from Pexels~\cite{pexels}. Subsequent content-tag filtering and human annotation yielded 63K high-quality clips characterized by high aesthetic, high resolution, diverse scenes, and stable motion. Using a structured captioning schema~\cite{seawead2025seaweed, wan2.1}, we annotated each clip with the Qwen2.5-VL~\cite{qwen2.5-vl}. The general statistics of dataset will be illustrated in the supplemented materials.

\begin{table*}[!thb]
\centering
\vspace{-0.9cm}
\caption{\textbf{Quantitative Comparison on Vbench} prompts from aesthetic and imaging quality dimensions. The best result is highlighted in \textbf{bold} and the second-best is \underline{underlined}.}
\vspace{-0.2cm}
\label{tab:vbench_dense}
\setlength{\tabcolsep}{5pt}
\renewcommand{\arraystretch}{0.84}
\small
\begin{tabular}{lcccccccccc}
\toprule
\multirow{3}{*}{Method} &
\multicolumn{2}{c}{\textit{Visual Quality}} &
\multicolumn{3}{c}{\textit{Human Preference}} &
\multicolumn{4}{c}{\textit{VideoAlign}} \\
\cmidrule(lr){2-3} \cmidrule(lr){4-6} \cmidrule(l){7-10}
 &
\makecell{Aesthetic Quality} &
\makecell{Imaging Quality} &
\makecell{HPS-v2} &
\makecell{PickScore} &
\makecell{Image Reward} &
\makecell{VQ} &
\makecell{MQ} &
\makecell{TA} &
\makecell{Overall} \\
\midrule
\textit{CogvideoX-2B:} \\
Baseline      & 0.6279 & 0.6589 & 0.2655 & \textbf{21.50} & \underline{0.6079} & 2.1430 & \underline{0.7741} & 4.8701 & 7.7871 \\
SFT           & 0.6293 & 0.6598 & \underline{0.2659} & \underline{21.47} & 0.5519 & \underline{2.2003} & 0.7496 & 4.6819 & 7.6318 \\
Vanilla DPO~\cite{videoalign}   & 0.6304 & 0.6598 & 0.2654 & 21.41 & 0.5972 & 2.1823 & \textbf{0.8067} & 4.7972 & 7.7862  \\
DenseDPO~\cite{densedpo} & \underline{0.6325} &
\underline{0.6606} &
0.2652 &
21.43 &
0.5884 &
2.1669 &
0.7675 &
\underline{4.8813} &
\underline{7.8157} \\
\cellcolor{mypink}\textbf{Ours} &\cellcolor{mypink}\textbf{0.6499} &\cellcolor{mypink}\textbf{0.7080} &\cellcolor{mypink}\textbf{0.2738} & \cellcolor{mypink}21.46 &\cellcolor{mypink}\textbf{0.6492} &\cellcolor{mypink}\textbf{2.2363} &\cellcolor{mypink}0.7173 &\cellcolor{mypink}\textbf{4.9031} &\cellcolor{mypink}\textbf{7.8568} \\
\midrule
\textit{CogvideoX-5B:} \\
Baseline      & 0.6110 & 0.6631 & 0.2692 & \underline{21.72} & 0.5957 & \underline{4.1696} & \underline{1.6005} & 3.9490 & \underline{9.7191} \\
SFT           & 0.6132 & 0.6860 & \underline{0.2728} & 21.58 & 0.5726 & 3.9869 & 1.4136 & 3.9619 & 9.3624 \\
Vanilla DPO~\cite{videoalign}   & 0.5953 & 0.6534 & 0.2658 & 21.56 & \underline{0.6012} & 4.0808 & 1.5498 & 3.9602 & 9.5910  \\
DenseDPO~\cite{densedpo} & \underline{0.6233} & \underline{0.6962} & 0.2674 & 21.67 & 0.5959 & 3.3251 & 1.2671 & \textbf{4.9804} & 9.5726 \\
\cellcolor{mypink}\textbf{Ours} & \cellcolor{mypink}\textbf{0.6274} & \cellcolor{mypink}\textbf{0.7107} & \cellcolor{mypink}\textbf{0.2782} & \cellcolor{mypink}\textbf{21.70} & \cellcolor{mypink}\textbf{0.6297} & \cellcolor{mypink}\textbf{4.5129} & \cellcolor{mypink}\textbf{1.6682} & \cellcolor{mypink}\underline{4.1118} & \cellcolor{mypink}\textbf{10.2930} \\
\midrule

\textit{Wan 2.1-1.3B:} \\
Baseline    & 0.6363 & 0.6296 & 0.2727 & 21.37 & 0.6874 & 1.9387 & \textbf{0.5468} & \underline{5.3444} & 7.8300  \\
SFT      & 0.6373 & 0.6342 & \underline{0.2730} & \underline{21.38} & \underline{0.7220} & 1.8779 & 0.5149 & 5.3355 & 7.7283     \\
Vanilla DPO~\cite{videoalign}   & 0.6353 & 0.6308 & 0.2654 & 21.37 & 0.5972 & 1.9437 & 0.5259 & 5.3383 & 7.8079   \\
DenseDPO~\cite{densedpo} & \underline{0.6375} &
\underline{0.6356} &
0.2728 &
21.37 &
0.6876 &
\underline{1.9519} &
0.5422 &
5.3431 &
\underline{7.8373} \\
\cellcolor{mypink}\textbf{Ours} &   \cellcolor{mypink}\textbf{0.6416} & \cellcolor{mypink}\textbf{0.6412} & \cellcolor{mypink}\textbf{0.2754} & \cellcolor{mypink}\textbf{21.42} & \cellcolor{mypink}\textbf{0.7297} &\cellcolor{mypink}\textbf{2.0652}   &\cellcolor{mypink}\underline{0.5465}  &\cellcolor{mypink}\textbf{5.3471}   &\cellcolor{mypink}\textbf{7.9588}   \\

\bottomrule
\end{tabular}
\end{table*}

\begin{table*}[!thb]
\centering
\caption{\textbf{Quantitative Comparison on VideoJAM benchmark}. The best result is highlighted in \textbf{bold} and the second-best is \underline{underlined}.}
\vspace{-0.2cm}
\label{tab:videojam_dense}
\setlength{\tabcolsep}{5.15pt}
\renewcommand{\arraystretch}{0.8}
\small
\begin{tabular}{lcccccccccc}
\toprule
\multirow{3}{*}{Method}  &
\multicolumn{2}{c}{\textit{Visual Quality}} &
\multicolumn{3}{c}{\textit{Human Preference}} &
\multicolumn{4}{c}{\textit{VideoAlign}} \\
\cmidrule(lr){2-3} \cmidrule(lr){4-6} \cmidrule(l){7-10}
 &
\makecell{Aesthetic Quality} &
\makecell{Imaging Quality} &
\makecell{HPS-v2} &
\makecell{PickScore} &
\makecell{Image Reward} &
\makecell{VQ} &
\makecell{MQ} &
\makecell{TA} &
\makecell{Overall} \\
\midrule
\textit{CogvideoX-2B:} \\
Baseline      & 0.5494 & 0.6327 & 0.2445 & 20.88 & 0.6407 & 1.7707 & \textbf{0.3849} & 5.3140 & 7.4696 \\
SFT           & \underline{0.5567} & \underline{0.6382} & \underline{0.2471} & \textbf{21.04} & \underline{0.6910} & 1.7966 & 0.3300 & 5.3368 & 7.4635 \\
Vanilla DPO~\cite{videoalign}   & 0.5482 & 0.6310 & 0.2443 & 20.96 & 0.6358 & \underline{1.8198} & 0.3446 & 5.3515 & 7.5160  \\
DenseDPO~\cite{densedpo}  & 0.5521 &
0.6334 &
0.2448 &
20.96 &
0.6501 &
1.8147 &
\underline{0.3568} &
\underline{5.3559} &
\underline{7.5214} \\
\cellcolor{mypink}\textbf{Ours} & \cellcolor{mypink}\textbf{0.5604} & \cellcolor{mypink}\textbf{0.7001} & \cellcolor{mypink}\textbf{0.2543} & \cellcolor{mypink}\underline{20.97} & \cellcolor{mypink}\textbf{0.7036} & \cellcolor{mypink}\textbf{1.8207} & \cellcolor{mypink}0.3134 & \cellcolor{mypink}\textbf{5.4054} & \cellcolor{mypink}\textbf{7.5397} \\
\midrule

\textit{CogvideoX-5B:} \\
Baseline      & 0.5631 & 0.6135 & 0.2421 & \underline{21.00} & 0.4805 & \underline{1.7597} & \underline{0.2987} & \underline{5.4428} & \underline{7.5012} \\
SFT           & \underline{0.5635} & 0.6166 & \underline{0.2445} & 20.99 & \underline{0.5485} & 1.7151 & 0.2771 & 5.4379 & 7.4301 \\
Vanilla DPO~\cite{videoalign}   & 0.5553 & 0.6148 & 0.2403 & 20.94 & 0.4996 & 1.7056 & 0.2785 & 5.3235 & 7.3076  \\
DenseDPO~\cite{densedpo} & 0.5614 & \underline{0.6171} & 0.2424 & 20.98 & 0.5188 & 1.7947 & 0.2640 & 5.3634 & 7.4220 \\
\cellcolor{mypink}\textbf{Ours} & \cellcolor{mypink}\textbf{0.5782} & \cellcolor{mypink}\textbf{0.6727} & \cellcolor{mypink}\textbf{0.2523} & \cellcolor{mypink}\textbf{21.03} & \cellcolor{mypink}\textbf{0.5707} & \cellcolor{mypink}\textbf{1.8785} & \cellcolor{mypink}\textbf{0.3190} & \cellcolor{mypink}\textbf{5.4451} & \cellcolor{mypink}\textbf{7.6424} \\
\midrule

\textit{Wan 2.1-1.3B:} \\
Baseline      & 0.5623 & 0.6021 & 0.2499 & 20.82 & 0.6292 & 1.3637 & \underline{0.1613} & \textbf{5.6295} & \underline{7.1545} \\
SFT           & \underline{0.5675} & 0.6003 & 0.2494 & 20.81 & 0.6302 & 1.3571 & 0.1555 & 5.5195 & 7.0321 \\
Vanilla DPO~\cite{videoalign}   & 0.5611 & \underline{0.6042} & \underline{0.2503} & \underline{20.83} & \underline{0.6496} & 1.3646 & 0.1357 & 5.5545 & 7.0548  \\
DenseDPO~\cite{densedpo} & 0.5622 &
0.6021 &
0.2501 &
20.82 &
0.6342 &
\underline{1.3657} &
0.1387 &
\underline{5.6156} &
7.1200 \\
\cellcolor{mypink}\textbf{Ours} & \cellcolor{mypink}\textbf{0.5698} & \cellcolor{mypink}\textbf{0.6467} & \cellcolor{mypink}\textbf{0.2533} & \cellcolor{mypink}\textbf{20.92} & \cellcolor{mypink}\textbf{0.6667} & \cellcolor{mypink}\textbf{1.7033} & \cellcolor{mypink}\textbf{0.2366} & \cellcolor{mypink}5.5450 & \cellcolor{mypink}\textbf{7.4849} \\
\bottomrule
\end{tabular}
\end{table*}

\begin{figure}[!thbp]
\centering
\vspace{-0.3cm}
\includegraphics[width=1.0\columnwidth]{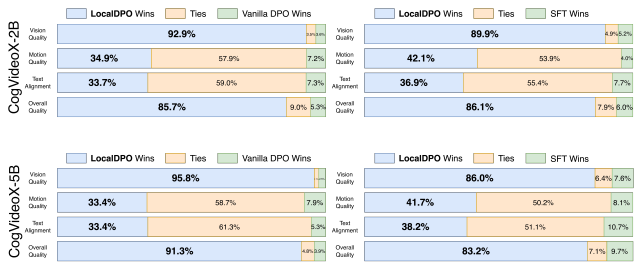} 
\vspace{-0.6cm}
\caption{Human evaluation of LocalDPO \emph{vs.} SFT and VanillaDPO. LocalDPO achieves the best results on all dimensions of human evaluation.}
\label{fig:userstudy}
\vspace{-0.5cm}
\end{figure}

\vspace{-0.1cm}
\subsection{Experimental Setup}

\noindent\textbf{Baselines and comparisons.}
To demonstrate the effectiveness of our method, we conduct extensive experiments on multiple DiT-based VDMs with varying parameter scales, including CogVideoX-2B~\citep{yang2024cogvideox}, CogVideoX-5B~\citep{yang2024cogvideox}, and Wan2.1-1.3B~\citep{wan2.1}.
We compare our method against:
(1) Baseline: The pretrained base model;
(2) SFT: The model finetuned on our 63K video dataset by LoRA;
(3) Vanilla DPO: Standard diffusion DPO approach using multi-sampled preference pairs. We generate three videos with different seeds per prompt, and rank these videos with a pretrained critic model~\cite{videoalign}.
(4) DenseDPO: An improved DPO method that considers different frames as the granularity of preference~\cite{densedpo}. We generate two videos with different seeds per prompt, and rank these videos in terms of frame-level with a pretrained critic model~\cite{videoalign}.
All experimental settings are fine-tuned using the same optimization protocol and the identical quantity of training data for fair comparison.

\noindent\textbf{Evaluation benchmarks.}
To evaluate the algorithm comprehensively, we utilize 165 VBench~\cite{vbench} prompts from aesthetic and imaging quality dimensions, along with the prompts from VideoJAM~\cite{chefer2025videojam}.
Each prompt is expanded by Qwen2.5-VL~\cite{qwen2.5-vl} to a richer and more detailed expression.
For a multi-faceted evaluation of the results, we employ three objective evaluation dimensions: 
(1) Visual Quality Metrics: aesthetic quality~\cite{schuhmann2021laion} and image quality~\cite{ke2021musiq} from VBench;
(2) Human Preference metrics: HPS-V2~\cite{hps-v2}, ImageReward~\cite{xu2023imagereward}, and PickScore~\cite{pickscore}; 
(3) Video Alignment metrics~\cite{videoalign}: Visual Quality (VQ), Motion Quality (MQ), Text Aligment (TA) and Overall Quality (Overall).

\noindent\textbf{Implementation details.}
For each real video, we generate random spatio-temporal masks using Bézier curves (as described in Sec.~\ref{sec:method}). During negative sample synthesis, we add noise at a random strength under $\alpha_l=0.75$ and $\alpha_h=0.95$, simulating moderate-to-strong corruption. We fine-tune models using LoRA~\cite{hu2022lora} with rank 64 on attention layers of DiT only, keeping the rest of the modules frozen. The total loss is $\mathcal{L}_{\text{total}} = \lambda_{\text{RA-DPO}} \mathcal{L}_{\text{RA-DPO}} + \lambda_{\text{DPO}} \mathcal{L}_{\text{GlobalDPO}} + \lambda_{\text{SFT}} \mathcal{L}_{\text{SFT}}$, with weights $\lambda_{\text{RA-DPO}} = 1.0$, $\lambda_{\text{DPO}} = 1.0$, and $\lambda_{\text{SFT}} = 0.1$. We train for 540 iterations with a batch size of 128 and adopt AdamW optimizer~\cite{loshchilovdecoupled} for our methods and other comparisons. During inference, we use 50 DDIM steps ~\cite{ddim} with classifier-free guidance scale 6.0.

\begin{table}[htbp]
\fontsize{7}{9.5}\selectfont
\begin{center}
\caption{Ablation on loss components. \checkmark indicates the used loss.}
\label{tab:ablation_loss}
\setlength{\tabcolsep}{0.4mm}{
\begin{tabular}{@{}lcccccccc@{}}
\toprule
\multirow{4}{*}{$\mathcal{L}_{\text{DPO}}$} & \multirow{4}{*}{$\mathcal{L}_{\text{SFT}}$} & \multirow{4}{*}{$\mathcal{L}_{\text{RA-DPO}}$} & \multicolumn{2}{c}{\textit{Visual Quality}} & \multicolumn{3}{c}{\textit{Human Preference}} & \textit{VideoAlign} \\ \cmidrule(r){4-5} \cmidrule(r){6-8} \cmidrule(r){9-9}
& & & \makecell{Aesthetic \\ Quality} & \makecell{Imaging \\ Quality} & HPS-v2 & PickScore & \makecell{Image \\ Reward} & Overall \\ \midrule
& & & 0.6279 & \underline{0.6589} & 0.2655 & \textbf{21.50} & \underline{0.6079} & 7.7871 \\
\checkmark & & & 0.6303 & 0.6522 & 0.2657 & 21.40 & 0.6075 & 7.7918 \\
\checkmark & \checkmark &  & \underline{0.6308} & 0.6514 & \underline{0.2659} & 21.41 & 0.6072  & \underline{7.8003} \\
\cellcolor{mypink}\checkmark& \cellcolor{mypink}\checkmark & \cellcolor{mypink}\checkmark & \cellcolor{mypink}\textbf{0.6499} & \cellcolor{mypink}\textbf{0.7080} & \cellcolor{mypink}\textbf{0.2738} & \cellcolor{mypink}\underline{21.46} & \cellcolor{mypink}\textbf{0.6492} & \cellcolor{mypink}\textbf{7.8568} \\
\bottomrule
\end{tabular}}
\end{center}
\vspace{-0.5cm}

\vspace{-0.35cm}
\end{table}

\begin{table}[htbp]
\fontsize{6.5}{9.5}\selectfont
\begin{center}
\caption{Ablation on positive and negative sample construction strategies. ``Vanilla win" and ``Vanilla lose" indicate the win and lose sample used in vanilla DPO. ``RA corruption" represents the region-aware corruption in our method.}
\label{tab:ablation_pos_neg}
\setlength{\tabcolsep}{0.4mm}{
\begin{tabular}{@{}cccccccc@{}}
\toprule
\multirow{3}{*}{\makecell{Positive \\ Sample}} & \multirow{3}{*}{\makecell{Negative \\ Sample}} & \multicolumn{2}{c}{\textit{Visual Quality}} & \multicolumn{3}{c}{\textit{Human Preference}} & \textit{VideoAlign} \\ \cmidrule(r){3-4} \cmidrule(r){5-7} \cmidrule(r){8-8}
& & \makecell{Aesthetic \\ Quality} & \makecell{Imaging \\ Quality} & HPS-v2 & PickScore & \makecell{Image \\ Reward} & Overall \\ \midrule
Vanilla win & Vanilla lose & \underline{0.6304} & \underline{0.6598} & 0.2654 & 21.41 & 0.5972 & \underline{7.7862} \\
Real Video & Vanilla lose & 0.6285 & 0.6577 & \underline{0.2656} & \underline{21.44} & \underline{0.6137}  & 7.7778 \\
\cellcolor{mypink}\textbf{Real Video} & \cellcolor{mypink}\textbf{RA corruption} & \cellcolor{mypink}\textbf{0.6499} & \cellcolor{mypink}\textbf{0.7080} & \cellcolor{mypink}\textbf{0.2738} & \cellcolor{mypink}\textbf{21.46} & \cellcolor{mypink}\textbf{0.6492} & \cellcolor{mypink}\textbf{7.8568} \\
\bottomrule
\end{tabular}}
\end{center}
\vspace{-0.5cm}
\vspace{-0.35cm}
\end{table}

\subsection{Main Results}

Tab.~\ref{tab:vbench_dense} and Tab.~\ref{tab:videojam_dense} provide quantitative comparisons of our method and other counterparts on three selected VDMs, evaluated on VBench and VideoJAM, respectively.
The experimental results demonstrate the superiority of our method on the vast majority of metrics. Notably, our method achieves a pronounced advantage over other methods in visual quality metrics (\ie, aesthetic quality and image quality score), indicating that our preference data construction strategy and region-aware preference learning effectively enhance the visual quality of the generated videos.

\subsection{User Study}
Following~\cite{videoalign}, we conduct a user study with 20 participants that evaluates different models along four dimensions, including Visual Quality (VQ), Motion Quality (MQ), Text Alignment (TA) and Overall Quality (Overall).
The evaluation adopts a pairwise format, assigning a “win or lose or tie” label on each dimension. We construct an evaluation set by randomly sampling 50 prompts from VBench~\cite{vbench} and upsample each prompt via~\cite{qwen2.5-vl}, enriching fine-grained details. For the CogVideoX models (2B and 5B), the assessment compares our method against supervised fine-tuning (SFT) and Vanilla DPO, respectively. As shown in \cref{fig:userstudy}, our method achieves significant improvements over the counterparts in all dimensions, especially in VQ and Overall quality, achieving an average win rate of 88.86\%. The detailed annotation requirements and additional results will be present in the supplementary material.

\begin{figure*}[!htbp]
\centering
\vspace{-0.9cm}
\includegraphics[width=1.0\textwidth]{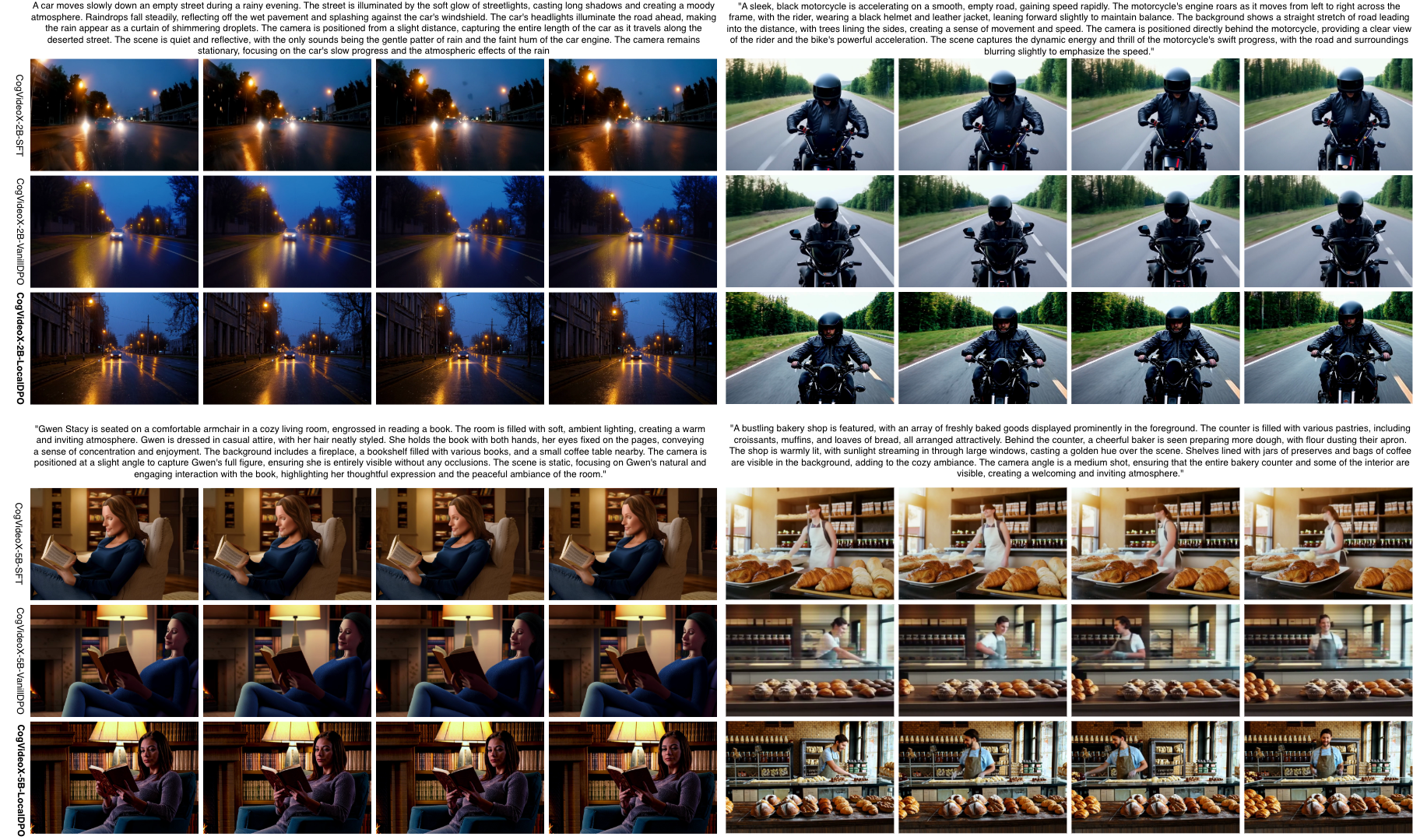} 
\vspace{-0.7cm}
\caption{Qualitative Comparison between SFT, Vanilla DPO and LocalDPO for CogVideoX models. Our LocalDPO generates rich textural details, plausible motion, higher aesthetic and fewer artifacts.}
\label{fig:qualitycomparison}
\end{figure*}

\subsection{Qualitative Comparison}
\cref{fig:qualitycomparison} illustrates the visual comparison generated by the main methods. We present our results on the third row in each comparative sample. The videos generated by our approach are markedly sharper and exhibit higher aesthetic quality. Owing to local detail preference optimization, they also contain richer details in both foreground subjects and background objects. Furthermore, our method preserves semantic alignment better, accurately realizing the specified style and target objects. In general, our method demonstrates an obvious subjective quality advantage compared to existing methods, and it improves objective metrics while simultaneously avoiding reward hacking. More comparative results will be presented in the supplementary material.

\begin{figure}[!htbp]
\centering
\vspace{-0.5cm}
\includegraphics[width=1.0\columnwidth]{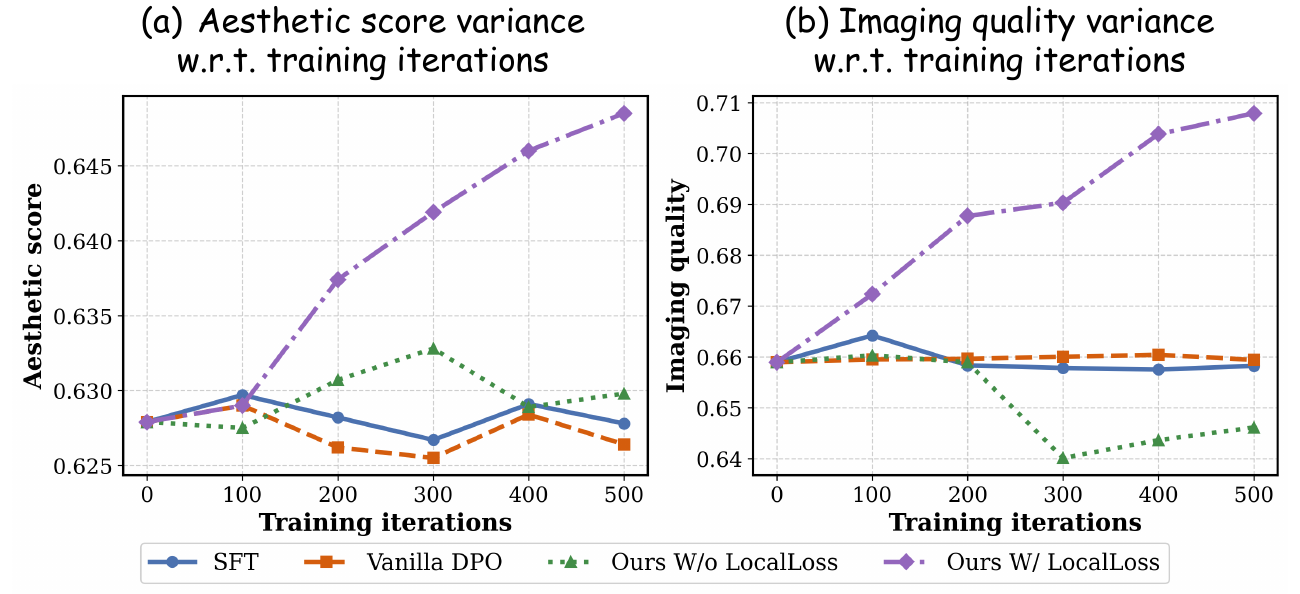} 
\vspace{-0.5cm}
\caption{Convergence of the models on aesthetic and image quality under different training iterations.}
\label{fig:localloss}
\vspace{-0.5cm}
\end{figure}

\subsection{Ablation Studies}

We conduct ablation studies on CogVideoX-2B, and we adopt prompts from Vbench aesthetic and imaging quality dimensions to validate key design choices:

\noindent\textbf{Impact of region-aware DPO loss.}
In the previous quantitative comparison experiments, our method is optimized using three loss terms jointly: region-aware DPO loss $\mathcal{L}_\text{RA-DPO}$, DPO loss $\mathcal{L}_\text{DPO}$, and SFT loss $\mathcal{L}_\text{SFT}$. Here, we investigate the impact of the region-aware DPO loss term on performance, with results presented in Tab.~\ref{tab:ablation_loss}.
As indicated in the comparison between the first three rows, the DPO loss $\mathcal{L}_\text{DPO}$ and SFT loss $\mathcal{L}_\text{SFT}$ bring subtle boost in visual quality and video align metrics. After introducing the region-aware DPO loss, we observe nearly all of the metrics improve significantly. We argue that the region-aware DPO loss emphasizes the regional impact in the DPO training, which helps the model to localize the divergence between the real videos and the corresponding parts in the locally corrupted negative samples, and the significantly regional difference in the visually consistent DPO pairs is beneficial for the model convergence. To further illustrate the impact of the loss terms on model performance, Fig.~\ref{fig:localloss} shows performance evolution of the model in terms of aesthetic and imaging quality scores during training under different loss combinations. It is clearly observable that after incorporating our region-aware DPO loss, the model performance improves more rapidly and achieves a higher upper bound.

\noindent\textbf{The effective of region-aware corruption}
A naive implementation in constructing the DPO training pairs without human labelling is to use the real world videos as the positive samples and the videos from the generative models as the negative samples. To validate the feasibility of this method, we construct the DPO training pairs with 63K real videos and the corresponding negative samples from vanilla DPO because these videos are generated from the model. 
We illustrate the results in Tab.~\ref{tab:ablation_pos_neg}. Due to the significant distribution divergence between the positive and the negative samples, the utilization of the real world videos and the generated videos as the DPO training pairs fails boosting the video generation ability. Especially in visual quality and video align metrics, such method cannot surpass the vanilla DPO. While our method adopts the region-aware corruption technique to construct the negative samples, which not only makes the positive samples and the negative samples more consistent in semantics, but also shrink the distribution gap between the positive and the negative samples. Compared with other counterparts, our method is more beneficial for the model to localize the subtle difference between the positive and the negative samples, which improves the video generation capabilities comprehensively.

%% file: sec/6_Conclusion.tex
\vspace{-0.2cm}
\section{Conclusion}
We presented LocalDPO, a fine-grained preference optimization framework for text-to-video diffusion models. By leveraging real videos as positive anchors and synthesizing localized negative samples through region-aware local corruption, our method constructs high-fidelity preference pairs without multi-sampling or human annotation. The proposed region-aware DPO loss enables region-specific alignment, while a hybrid training objective ensures global coherence and stability. Extensive experiments on CogVideoX models and Wan2.1 show consistent improvements over existing post-training strategies in both automatic metrics and human evaluations. 
\\

\noindent\textbf{Acknowledgment}
This work was supported in part by the National Key R$\&$D Program of China under Grant No. 2022YFA1004103. 

%% file: sec/X_suppl.tex

\section{3D Mask Generation Algorithm for Negative Videos Generation}
\label{sec:3DMG}
As described in the main text, the negative samples in our LocalDPO are obtained by applying localized corruption to real videos.
To select the regions to corrupt, we propose a Bézier curve–based localized region corruption algorithm, which is shown in Alg~.\ref{alg:contour_to_mask}.

Generally, our mask generation strategy is grounded in the principle of structured randomness: rather than using arbitrary pixel-level noise or simplistic geometric primitives (e.g., rectangles or ellipses), we generate temporally plausible occlusions by modeling them as smooth, closed contours with controllable irregularity. The core idea is to first construct a compact, non-convex shape through stochastic corruption of a circular template, then embed it at a random location within the video frame. This ensures that the resulting masks mimic real-world occluders—such as objects or foreground entities—that are typically compact, connected, and exhibit organic boundaries. By decoupling shape generation (via Bézier-spline-based contours) from spatial placement, our method offers both diversity and physical plausibility for region-aware video corruption.
Specifically,  \(k\) anchor points are sampled on a perturbed circle in polar coordinates, where the radial distance of each point is uniformly randomized within \([1-\rho, 1+\rho]\) to introduce shape irregularity. The resulting point set is then normalized by its axis-aligned bounding box and rescaled to a prescribed proposal region of size \(h \times w\). This resized shape is randomly translated within a full video frame of size \(H \times W\) by sampling a valid top-left offset.
Then smoothness is enforced by connecting consecutive anchor points with cubic Bézier curves, where control points are placed along the chord directions with a fixed scaling factor \(\alpha\). Finally, the closed spline is rasterized onto the \(H \times W\) grid to produce a binary mask \(R \in \{0,1\}^{H \times W}\), where pixels inside or on the contour are set to 1 and others to 0. In practice, for each sample, \( k \) is randomly sampled from the range \( 6 \) to \( 8 \), \( \rho \) is randomly sampled from the interval \( [0.6, 0.8] \), \( \alpha \) is randomly set within \( [0.2, 0.4] \), and \( h \) and \( w \) are randomly sampled from \( [H/3, H] \) and \( [W/3, W] \), respectively.

\begin{algorithm}[t]
\caption{Generate Binary Mask from Random Closed Contour}
\label{alg:contour_to_mask}
\begin{algorithmic}[1]
\Require Number of primary vertices $k \in \mathbb{Z}_{+}$, corruption ratio $\rho \in (0,1)$, 
         proposal region size $(h, w)$, video frame size $(H, W)$
\Ensure Binary mask $R \in \{0,1\}^{H \times W}$

\State \textcolor{blue}{// Step 1: Sample anchor points on a perturbed circle}
\For{$j = 0$ to $k-1$}
    \State Compute base angle: $\phi_j \gets \frac{2\pi j}{k}$
    \State Sample radial offset: $r_j \gets 1 - \rho + 2\rho \cdot u_j$, where $u_j \sim \mathcal{U}(0,1)$
    \State Set anchor point: $\mathbf{a}_j \gets r_j \cdot \big( \cos \phi_j,\ \sin \phi_j \big)^\top$
\EndFor

\State \textcolor{blue}{// Step 2: Compute axis-aligned bounding box and normalize to (h, w)}
\State $x_{\min} \gets \min_j a_j^{(x)}, \quad x_{\max} \gets \max_j a_j^{(x)}$
\State $y_{\min} \gets \min_j a_j^{(y)}, \quad y_{\max} \gets \max_j a_j^{(y)}$
\State $w_{\text{bbox}} \gets x_{\max} - x_{\min}, \quad h_{\text{bbox}} \gets y_{\max} - y_{\min}$
\For{$j = 0$ to $k-1$}
    \State $a_j^{(x)} \gets \frac{a_j^{(x)} - x_{\min}}{w_{\text{bbox}}} \cdot w$
    \State $a_j^{(y)} \gets \frac{a_j^{(y)} - y_{\min}}{h_{\text{bbox}}} \cdot h$
\EndFor

\State \textcolor{blue}{// Step 3: Randomly place the resized shape in the (H, W) canvas}
\State Sample top-left corner: $x_0 \sim \mathcal{U}\big(0,\ H - h\big), \quad y_0 \sim \mathcal{U}\big(0,\ W - w\big)$
\For{$j = 0$ to $k-1$}
    \State $a_j^{(x)} \gets a_j^{(x)} + y_0$  \Comment{image x-axis is horizontal (column)}
    \State $a_j^{(y)} \gets a_j^{(y)} + x_0$  \Comment{image y-axis is vertical (row)}
\EndFor

\State \textcolor{blue}{// Step 4: Construct cubic Bézier segments between consecutive anchors}

\State Let $\mathbf{a}_{k} \equiv \mathbf{a}_0$ (cyclic indexing)
\For{$j = 0$ to $k-1$}
    \State Compute direction vectors:
        $\mathbf{d}_{j+1} = \mathbf{a}_{j+1} - \mathbf{a}_j$
    \State Place first control point near $\mathbf{a}_j$ along outgoing direction:
        $\mathbf{c}_j^{(1)} \gets \mathbf{a}_j + \alpha \cdot \mathbf{d}_{j+1}$
    \State Place second control point near $\mathbf{a}_{j+1}$ along incoming direction:
        $\mathbf{c}_j^{(2)} \gets \mathbf{a}_{j+1} - \alpha \cdot \mathbf{d}_{j+1}$
    \State \textcolor{blue}{// $\alpha > 0$ controls curve smoothness (e.g., $\alpha = 1/3$)}
\EndFor

\State \textcolor{blue}{// Step 5: Form closed spline and rasterize}
\State Define closed contour $\mathcal{C}$ as the union of $k$ cubic Bézier curves,
    each parameterized by $\big( \mathbf{a}_j,\ \mathbf{c}_j^{(1)},\ \mathbf{c}_j^{(2)},\ \mathbf{a}_{j+1} \big)$
\State Rasterize $\mathcal{C}$ onto a 2D grid of size $(H, W)$:
    set pixel $(i,j) = 1$ if it lies inside or on $\mathcal{C}$, else $0$
\State \Return binary mask $R$
\end{algorithmic}
\end{algorithm}

\section{General Statistics of the Real Videos Dataset}
\subsection{Overview}

Following the data-construction pipeline~\cite{wang2025koala, ju2024miradata, wan2.1} and the filtering protocols~\cite{tschannen2025siglip, wei2024got, soucek2024transnet, schuhmann2021laion, raft, ke2021musiq, dover}, we curate a large dataset containing initial video clips from Pexels~\cite{pexels}. Subsequent content-tag filtering and human annotation yield 63K high-quality clips characterized by high aesthetic, high resolution, diverse scenes, and stable motion. Using a structured captioning schema~\cite{seawead2025seaweed, wan2.1}, we annotate each clip with Qwen2.5-VL~\cite{qwen2.5-vl}.

\begin{figure*}[!htbp]
\centering

\includegraphics[width=1.0\textwidth]{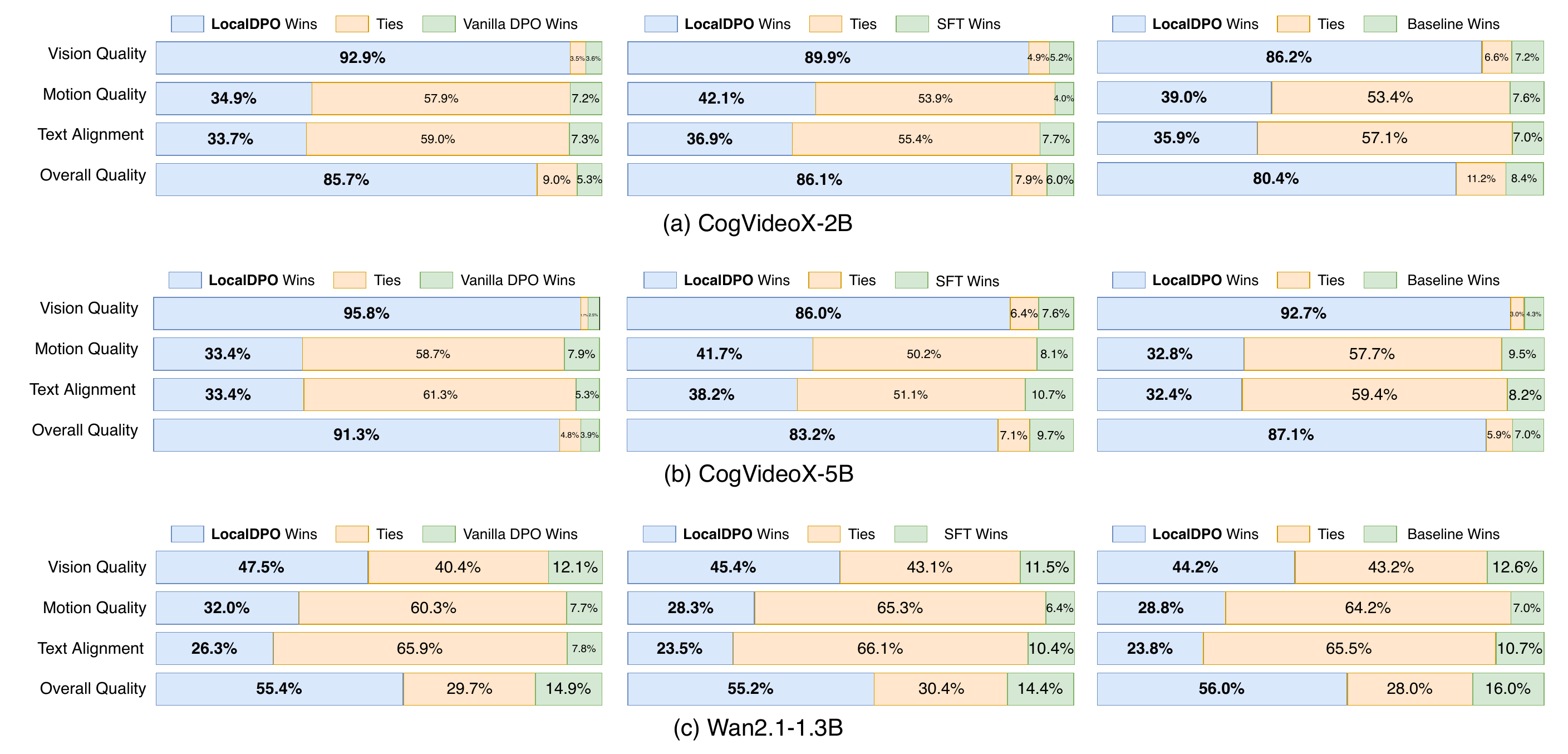} 

\caption{Human evaluation of LocalDPO~\emph{vs.}~Baseline, SFT and Vanilla DPO on CogvideoX-2B~\cite{yang2024cogvideox}, CogvideoX-5B~\cite{yang2024cogvideox} and Wan2.1-1.3B~\cite{wan2.1}. LocalDPO achieves the best results on all dimensions of human evaluation.}
\label{fig:qualitycomparison_wan}
\end{figure*}

\subsection{Preprocessing Pipeline of Real-World Videos}
To facilitate rigorous evaluation of video generation models, we construct a large-scale, high-quality video dataset from a real-world source. This section details the systematic pipeline for its collection, filtering, and annotation.

\subsubsection{\textbf{Data Source}}
Our primary data source is from Pexels ~\cite{pexels}, an extensive repository of royalty-free stock videos. We choose Pexels for its vast diversity in subjects, scenes, and motion patterns, as well as its high technical quality (HD, 4K formats). Our selection process aims to create a challenging and varied dataset using a keyword-based search strategy.

\subsubsection{\textbf{Video Selection Criteria}}
Our selection process is guided by the objective of creating a dataset that is both diverse and challenging. We employ a keyword-based search strategy with the following criteria:
\begin{itemize}
    \item[] \textbf{Scene Diversity:} A mix of environments, including keywords like \textit{``indoor," ``outdoor," ``city,"} and \textit{``nature."}
    \item[] \textbf{Motion Complexity:} A spectrum from static shots to highly dynamic content, using keywords such as \textit{``walking," ``running,"} and \textit{``slow motion."}
    \item[] \textbf{Subject Matter:} A balance of subjects including \textit{``people," ``animals," ``vehicles,"} and \textit{``objects."}
    \item[] \textbf{Technical Quality:} Only videos with a minimum resolution of 1080p and standard frame rates (24-60 FPS) are considered.
\end{itemize}

\subsubsection{\textbf{Data Filtering and Quality Assurance}}
To ensure a high standard of quality, every video is passed through a multi-stage automated filtering pipeline. Videos are discarded if they fail to meet predefined quality thresholds, assessed using the following state-of-the-art methods:

\begin{itemize}[label={}, leftmargin=*]
    \item \textbf{Technical Quality:}
    The DOVER model~\cite{dover} is used to assess a wide range of technical artifacts, providing a robust measure of overall fidelity.

    \item \textbf{Clarity:}
    The MUSIQ model~\cite{ke2021musiq}, a no-reference image quality assessor, is employed to ensure high sharpness and filter out blurry content.

    \item \textbf{Aesthetics:}
    A pre-trained aesthetic scoring model~\cite{schuhmann2021laion} is utilized to evaluate the perceptual and artistic appeal of each frame.

    \item \textbf{Motion Smoothness:}
    The ``vmafmotion" filter from FFmpeg and~\cite{raft} are applied to quantify motion, ensuring camera stability and removing clips with excessively shaky movements.

    \item \textbf{Text and Watermark Detection:}
    An OCR-based approach combining SigLIP~\cite{tschannen2025siglip} for region proposal and GOT~\cite{wei2024got} for text recognition are used to detect and remove on-screen watermarks.

    \item \textbf{Shot Integrity:}
    The TransNetV2 model~\cite{soucek2024transnet} is utilized to identify and exclude videos containing scene transitions, ensuring each video clip contains a single, continuous shot.
\end{itemize}

\subsubsection{\textbf{Caption Annotation Pipeline}}
We generate descriptive captions for each video using a state-of-the-art Video Large Language Model (VLLM), Qwen2.5-VL-7B~\cite{qwen2.5-vl}. To elicit professional-grade descriptions, we design a detailed prompt that instructs the model to analyze key visual elements (subject, motion, scene) and adopt specific narrative constraints, such as describing camera work from a photographer's perspective and avoiding phrases like ``the video shows." The prompt is presented as follows:
\begin{quote}
\textit{“Please describe the subject, motion, background, scene, camera motion, and style of this video in detail. Describe the camera motion as a professional photographer. If there are multiple subjects, clearly describe their spatial relationship. Do not use "the video" or "this video" as the subject of the sentence; directly start the sentence with the subject in the video. Keep the description clear and to the point, avoiding unnecessary details or repetition. Provide a coherent description without breaking it into sections or lists.”}
\end{quote}

\subsubsection{\textbf{Dataset Statistics}}
\begin{figure}[h!]
\centering
\includegraphics[width=1.0\columnwidth]{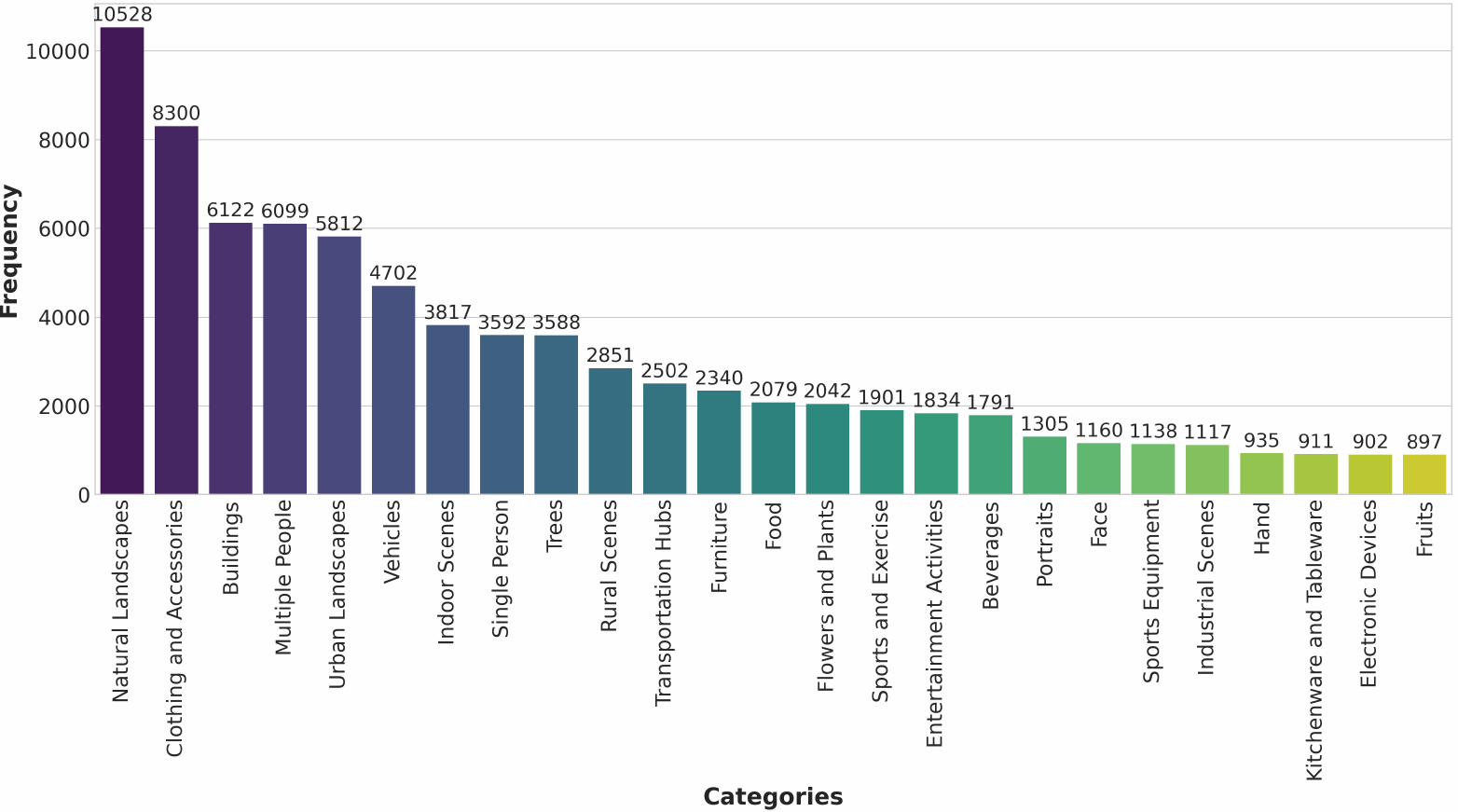} 
\caption{Category Distribution of the constructed video dataset.}
\label{fig:dataset_dist}
\end{figure}

Our pipeline results in a dataset including \textbf{63K diverse video clips}. The technical specifications and thematic distribution are presented below. Tab.~\ref{tab:dataset_specs} summarizes the key metrics of the dataset, while \cref{fig:dataset_dist} visualizes the category distribution, confirming a well-balanced composition for robust evaluation.

\begin{table}[h!]
\centering
\caption{Statistics of the curated data on key attributes.}
\label{tab:dataset_specs}
\begin{tabular}{ll}
\hline
\textbf{Metric}             & \textbf{Value / Range}         \\ \hline
Total Videos          & 63K                   \\
Resolution            & 1080p, 4K             \\
Frame Rate (FPS)      & 24-60        \\
Average Duration (s)  & 9.5                  \\ \hline
\end{tabular}
\end{table}

\begin{figure*}[!htbp]
\centering
\includegraphics[width=0.9\textwidth]{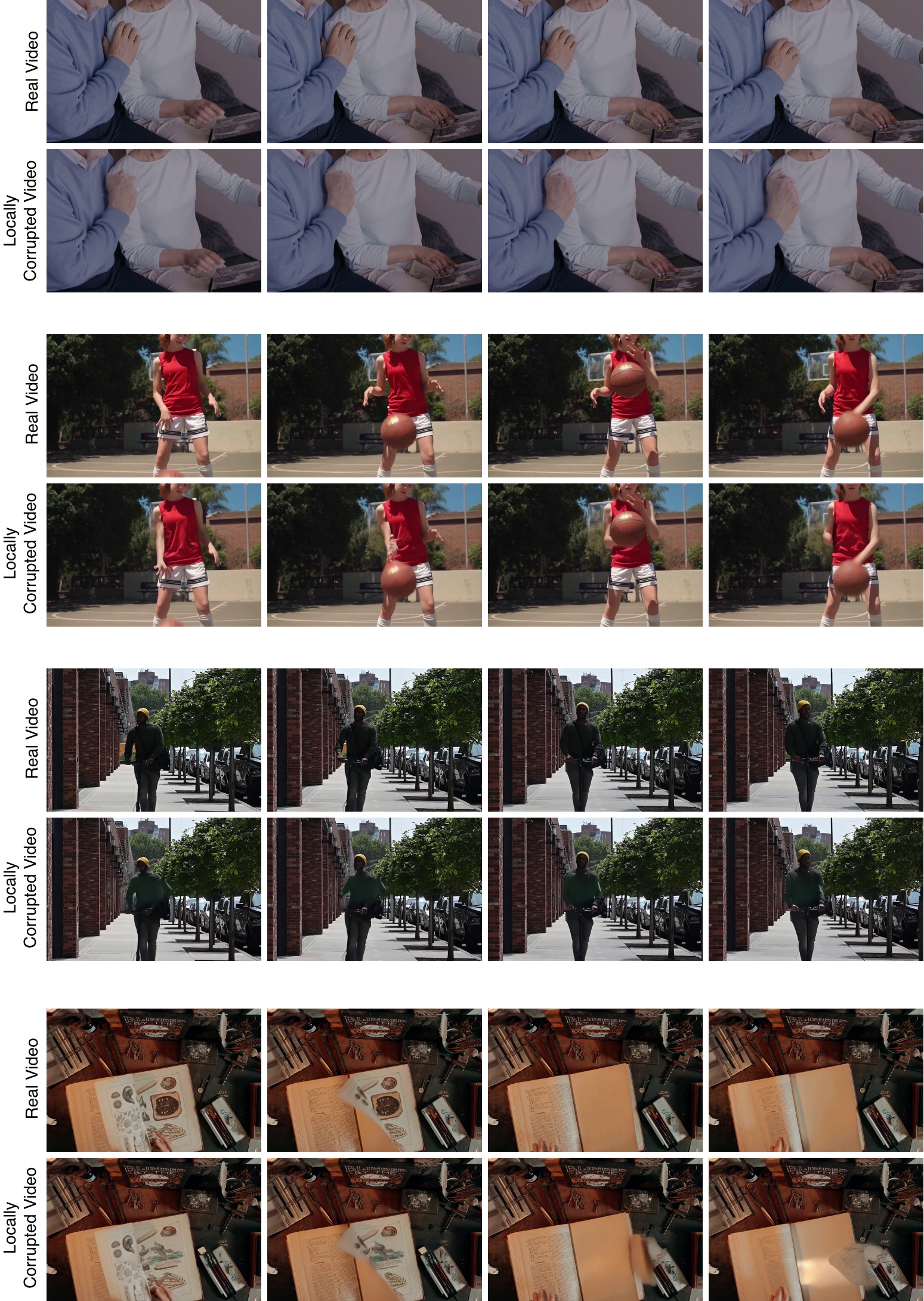} 
\caption{Visualization of generated locally corrupted videos.}
\label{fig:vis_neg}
\end{figure*}

\begin{figure*}[!htbp]
\centering
\includegraphics[width=1.0\textwidth]{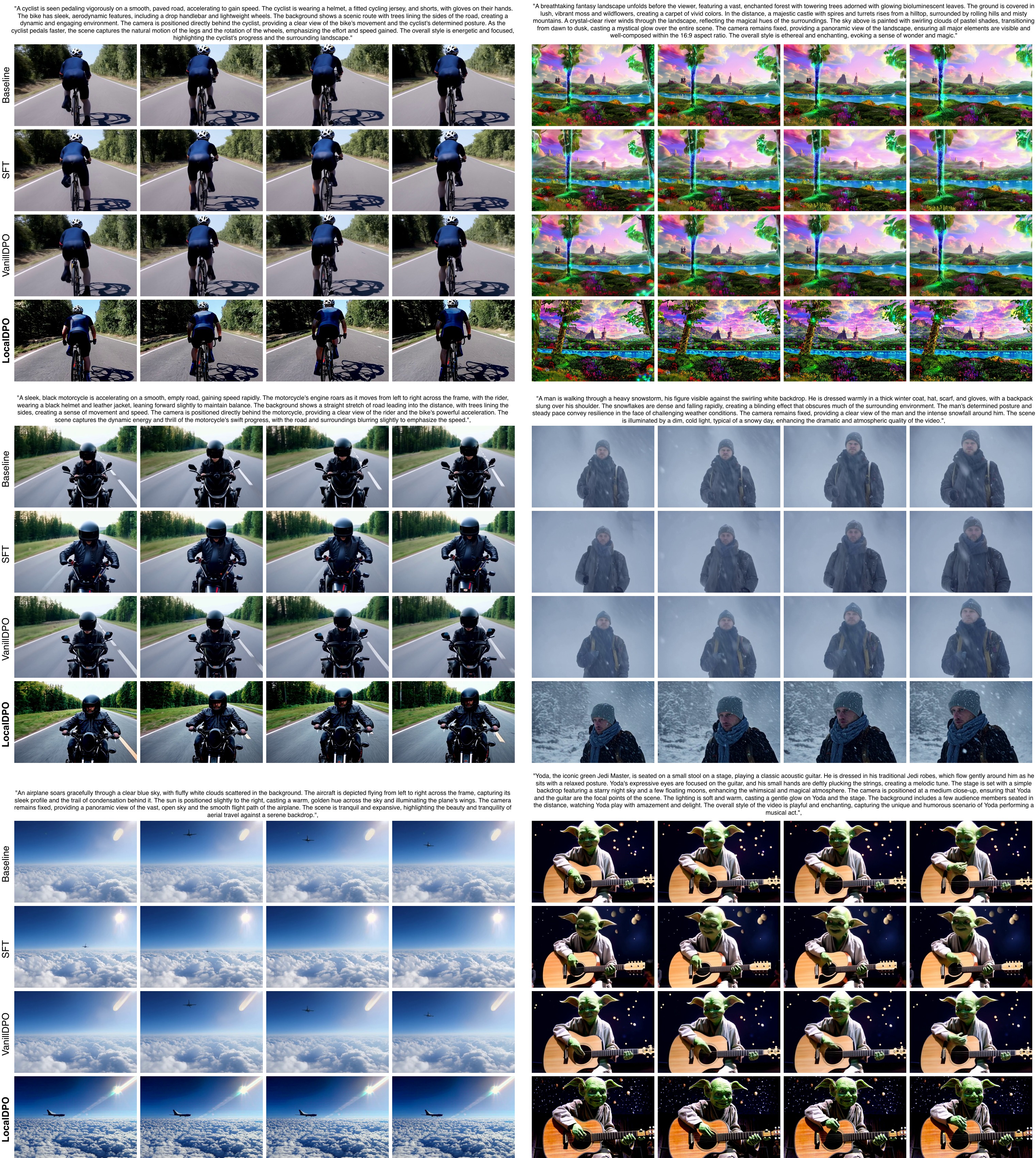} 
\caption{Visualization of LocalDPO \emph{vs.} Baseline, SFT and VanillaDPO on CogvideoX-2B.}
\label{fig:vis_cog2b}
\end{figure*}

\begin{figure*}[!htbp]
\centering
\includegraphics[width=1.0\textwidth]{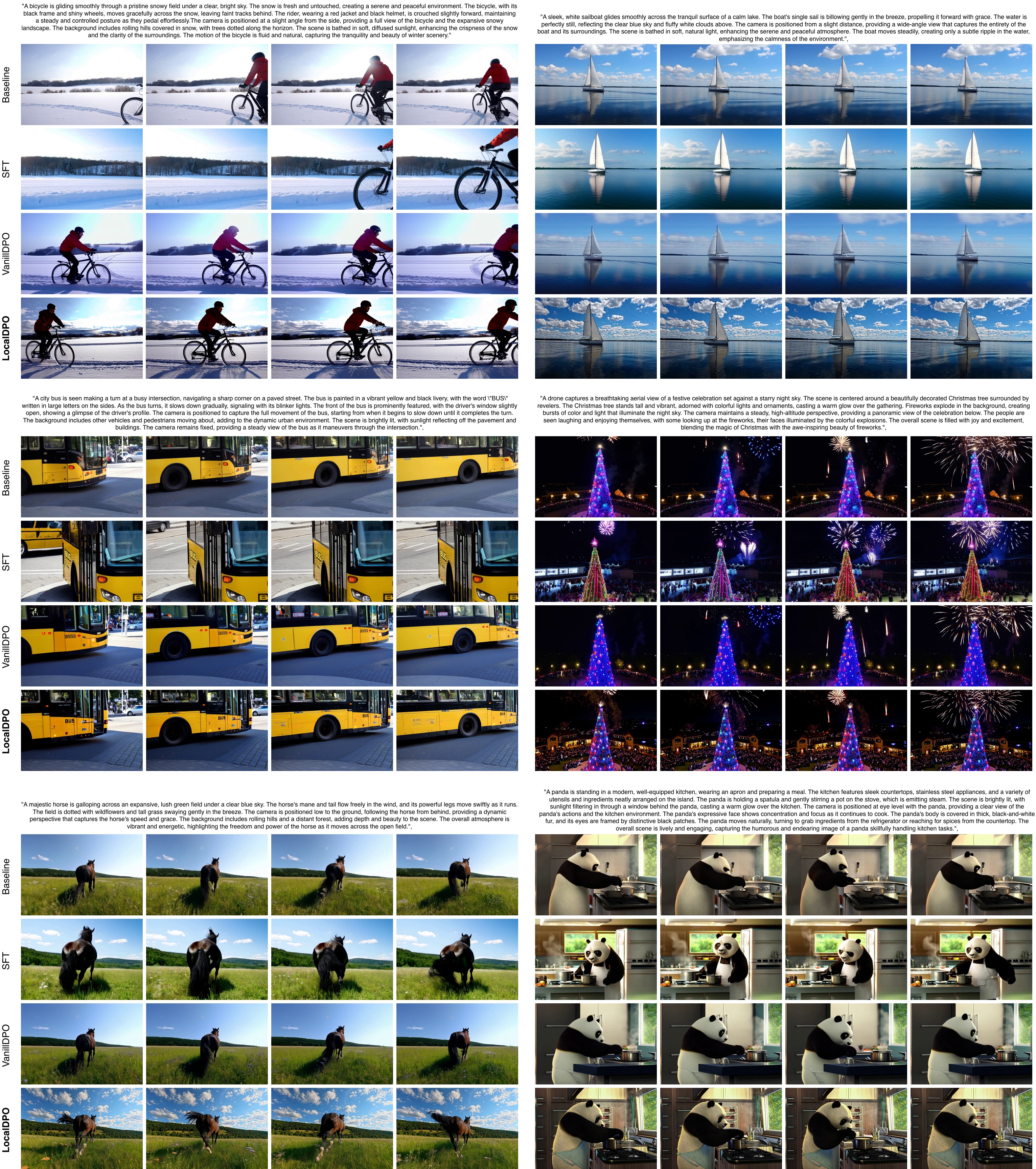} 
\caption{Visualization of LocalDPO \emph{vs.} Baseline, SFT and VanillaDPO on CogvideoX-5B.}
\label{fig:vis_cog5b}
\end{figure*}

\begin{figure*}[!htbp]
\centering
\includegraphics[width=1.0\textwidth]{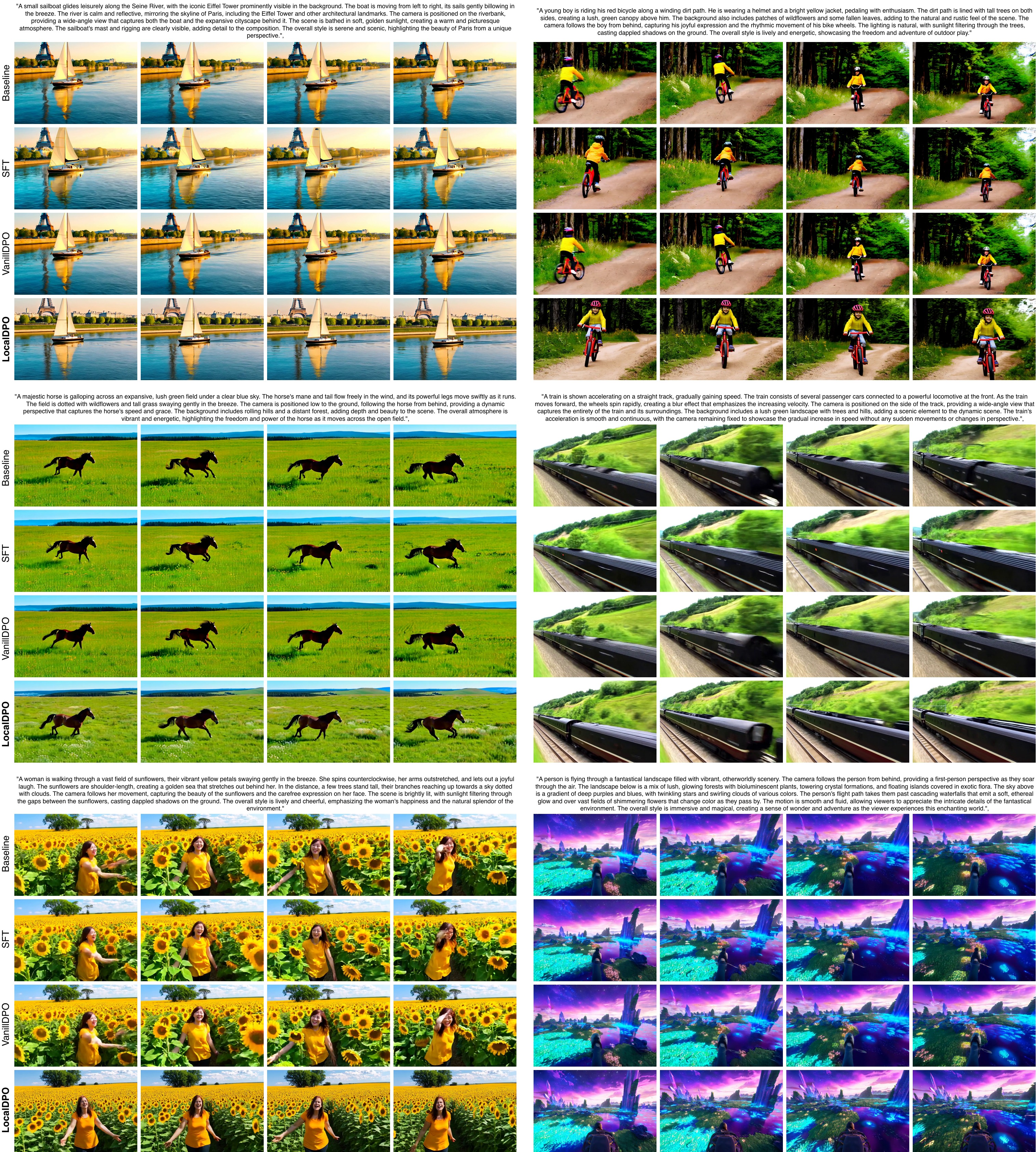} 
\caption{Visualization of LocalDPO \emph{vs.} Baseline, SFT and VanillaDPO on Wan2.1-1.3B.}
\label{fig:vis_wan}
\end{figure*}

\section{Additional Human Evaluation}
We present additional human evaluation results for CogVideoX-2B~\cite{yang2024cogvideox}, CogVideoX-5B~\cite{yang2024cogvideox}, and Wan2.1-1.3B~\cite{wan2.1} across four evaluation dimensions: Visual Quality (VQ), Motion Quality (MQ), Text Alignment (TA), and Overall Quality in Fig.~\ref{fig:qualitycomparison_wan}.
We compare our method with the baseline model, Supervised Fine-Tuning (SFT) and Vanilla DPO for comprehensive human evaluation.. 
Generally, the voting distributions consistently indicate that our method is preferred by a larger proportion of participants than either method in all four dimensions, further corroborating the superiority of our approach in human perceptual evaluation.



\section{Visualization of the LocalDPO training pairs}
In our LocalDPO, negative samples are constructed by applying localized corruption to the positive samples (i.e., real videos). In this subsection, we visualize the perturbed negative samples alongside their corresponding original videos (positive samples), as shown in Fig.~\ref{fig:vis_neg}. It is clearly observable that the perturbed regions often exhibit artifacts, distortions, or blurriness compared to the authentic video content, thereby forming reasonable training pairs that encode fine-grained, local-level preferences. Moreover, these imperfections precisely reflect the current limitations of pre-trained video generation models; consequently, training with such negative samples provides explicit feedback that effectively guides the model toward gradual improvement.

\section{Limitations and Future Work}
Our current approach generates spatio-temporal masks via random Bézier curves, which ensures diversity in corrupted regions but may lacks semantic awareness. 
Specifically, the corruptions are not tailored to particular object categories or semantic parts (e.g., faces, hands, or vehicles), potentially overlooking critical regions where quality degradation most affects user perception. 
As a result, the preference signal may be less effective for improving generation fidelity of specific object classes. 

In future work, we will incorporate vision foundation models, such as Grounding DINO~\cite{liu2024grounding} for object detection and SAM~\cite{kirillov2023segment,ravisam} for segmentation, to guide mask placement towards semantically meaningful regions. 
This would enable targeted refinement of object-level realism and controllability in text-to-video generation.

\section{More Qualitative Comparisons}
We present additional visual comparisons between our method and other methods, including the baseline, SFT, and vanilla DPO. Fig~\ref{fig:vis_cog2b}, Fig~\ref{fig:vis_cog5b}, and Fig~\ref{fig:vis_wan} show comparisons based on CogVideoX-2B, CogVideoX-5B, and Wan2.1-1.3B, respectively. Clearly, our LocalDPO generates videos with higher visual quality, better captures fine-grained details of the subject, and more faithfully adheres to the appearance.
These consistency results strongly demonstrate the effectiveness of our LocalDPO, particularly in enhancing video quality and preserving subject details.


%% file: main.bib
@String(PAMI = {IEEE Trans. Pattern Anal. Mach. Intell.})

@String(IJCV = {Int. J. Comput. Vis.})

@String(CVPR= {IEEE Conf. Comput. Vis. Pattern Recog.})

@String(ICCV= {Int. Conf. Comput. Vis.})

@String(ECCV= {Eur. Conf. Comput. Vis.})

@String(ACMMM= {ACM Int. Conf. Multimedia})

@String(ICLR = {Int. Conf. Learn. Represent.})

@String(PAMI  = {IEEE TPAMI})

@String(IJCV  = {IJCV})

@String(CVPR  = {CVPR})

@String(ICCV  = {ICCV})

@String(ECCV  = {ECCV})

@String(ACMMM = {ACM MM})

@String(ICLR  = {ICLR})

@article{croitoru2023diffusion,
  title={Diffusion models in vision: A survey},
  author={Croitoru, Florinel-Alin and Hondru, Vlad and Ionescu, Radu Tudor and Shah, Mubarak},
  journal={PAMI},
  volume={45},
  number={9},
  pages={10850--10869},
  year={2023},
  publisher={Ieee}
}

@article{yang2023diffusion,
  title={Diffusion models: A comprehensive survey of methods and applications},
  author={Yang, Ling and Zhang, Zhilong and Song, Yang and Hong, Shenda and Xu, Runsheng and Zhao, Yue and Zhang, Wentao and Cui, Bin and Yang, Ming-Hsuan},
  journal={ACM computing surveys},
  volume={56},
  number={4},
  pages={1--39},
  year={2023},
  publisher={ACM New York, NY, USA}
}

@article{ho2020denoising,
  title={Denoising diffusion probabilistic models},
  author={Ho, Jonathan and Jain, Ajay and Abbeel, Pieter},
  journal={NeurIPS},
  volume={33},
  pages={6840--6851},
  year={2020}
}

@article{ho2022video,
  title={Video diffusion models},
  author={Ho, Jonathan and Salimans, Tim and Gritsenko, Alexey and Chan, William and Norouzi, Mohammad and Fleet, David J},
  journal={NeurIPS},
  volume={35},
  pages={8633--8646},
  year={2022}
}

@article{blattmann2023stable,
  title={Stable video diffusion: Scaling latent video diffusion models to large datasets},
  author={Blattmann, Andreas and Dockhorn, Tim and Kulal, Sumith and Mendelevitch, Daniel and Kilian, Maciej and Lorenz, Dominik and Levi, Yam and English, Zion and Voleti, Vikram and Letts, Adam and others},
  journal={arXiv preprint arXiv:2311.15127},
  year={2023}
}

@inproceedings{yang2024cogvideox,
  title={CogVideoX: Text-to-Video Diffusion Models with An Expert Transformer},
  author={Yang, Zhuoyi and Teng, Jiayan and Zheng, Wendi and Ding, Ming and Huang, Shiyu and Xu, Jiazheng and Yang, Yuanming and Hong, Wenyi and Zhang, Xiaohan and Feng, Guanyu and others},
  booktitle={ICLR},
  year={2025}
}

@article{xing2024survey,
  title={A survey on video diffusion models},
  author={Xing, Zhen and Feng, Qijun and Chen, Haoran and Dai, Qi and Hu, Han and Xu, Hang and Wu, Zuxuan and Jiang, Yu-Gang},
  journal={ACM Computing Surveys},
  volume={57},
  number={2},
  pages={1--42},
  year={2024},
  publisher={ACM New York, NY}
}

@article{schulman2017proximal,
  title={Proximal policy optimization algorithms},
  author={Schulman, John and Wolski, Filip and Dhariwal, Prafulla and Radford, Alec and Klimov, Oleg},
  journal={arXiv preprint arXiv:1707.06347},
  year={2017}
}

@article{rafailov2023direct,
  title={Direct preference optimization: Your language model is secretly a reward model},
  author={Rafailov, Rafael and Sharma, Archit and Mitchell, Eric and Manning, Christopher D and Ermon, Stefano and Finn, Chelsea},
  journal={NeurIPS},
  volume={36},
  pages={53728--53741},
  year={2023}
}

@article{densedpo,
  title={DenseDPO: Fine-Grained Temporal Preference Optimization for Video Diffusion Models},
  author={Wu, Ziyi and Kag, Anil and Skorokhodov, Ivan and Menapace, Willi and Mirzaei, Ashkan and Gilitschenski, Igor and Tulyakov, Sergey and Siarohin, Aliaksandr},
  journal={NeurIPS},
  year={2025}
}

@article{videoalign,
  title={Improving video generation with human feedback},
  author={Liu, Jie and Liu, Gongye and Liang, Jiajun and Yuan, Ziyang and Liu, Xiaokun and Zheng, Mingwu and Wu, Xiele and Wang, Qiulin and Xia, Menghan and Wang, Xintao and others},
  journal={NeurIPS},
  year={2025}
}

@inproceedings{videodpo,
  title={Videodpo: Omni-preference alignment for video diffusion generation},
  author={Liu, Runtao and Wu, Haoyu and Zheng, Ziqiang and Wei, Chen and He, Yingqing and Pi, Renjie and Chen, Qifeng},
  booktitle={CVPR},
  pages={8009--8019},
  year={2025}
}

@inproceedings{personalvideo,
  title={Personalvideo: High id-fidelity video customization without dynamic and semantic degradation},
  author={Li, Hengjia and Qiu, Haonan and Zhang, Shiwei and Wang, Xiang and Wei, Yujie and Li, Zekun and Zhang, Yingya and Wu, Boxi and Cai, Deng},
  booktitle={ICCV},
  pages={19406--19416},
  year={2025}
}

@article{lift,
  title={Lift: Leveraging human feedback for text-to-video model alignment},
  author={Wang, Yibin and Tan, Zhiyu and Wang, Junyan and Yang, Xiaomeng and Jin, Cheng and Li, Hao},
  journal={arXiv preprint arXiv:2412.04814},
  year={2024}
}

@inproceedings{dover,
  title={Exploring video quality assessment on user generated contents from aesthetic and technical perspectives},
  author={Wu, Haoning and Zhang, Erli and Liao, Liang and Chen, Chaofeng and Hou, Jingwen and Wang, Annan and Sun, Wenxiu and Yan, Qiong and Lin, Weisi},
  booktitle={ICCV},
  pages={20144--20154},
  year={2023}
}

@inproceedings{ke2021musiq,
  title={Musiq: Multi-scale image quality transformer},
  author={Ke, Junjie and Wang, Qifei and Wang, Yilin and Milanfar, Peyman and Yang, Feng},
  booktitle={ICCV},
  pages={5148--5157},
  year={2021}
}

@article{tschannen2025siglip,
  title={Siglip 2: Multilingual vision-language encoders with improved semantic understanding, localization, and dense features},
  author={Tschannen, Michael and Gritsenko, Alexey and Wang, Xiao and Naeem, Muhammad Ferjad and Alabdulmohsin, Ibrahim and Parthasarathy, Nikhil and Evans, Talfan and Beyer, Lucas and Xia, Ye and Mustafa, Basil and others},
  journal={arXiv preprint arXiv:2502.14786},
  year={2025}
}

@article{wei2024got,
  title={General ocr theory: Towards ocr-2.0 via a unified end-to-end model},
  author={Wei, Haoran and Liu, Chenglong and Chen, Jinyue and Wang, Jia and Kong, Lingyu and Xu, Yanming and Ge, Zheng and Zhao, Liang and Sun, Jianjian and Peng, Yuang and others},
  journal={arXiv preprint arXiv:2409.01704},
  year={2024}
}

@inproceedings{soucek2024transnet,
  title={Transnet v2: An effective deep network architecture for fast shot transition detection},
  author={Soucek, Tom{\'a}s and Lokoc, Jakub},
  booktitle={ACMMM},
  pages={11218--11221},
  year={2024}
}

@article{qwen2.5-vl,
  title={Qwen2. 5-vl technical report},
  author={Bai, Shuai and Chen, Keqin and Liu, Xuejing and Wang, Jialin and Ge, Wenbin and Song, Sibo and Dang, Kai and Wang, Peng and Wang, Shijie and Tang, Jun and others},
  journal={arXiv preprint arXiv:2502.13923},
  year={2025}
}

@InProceedings{vbench,
    author    = {Huang, Ziqi and He, Yinan and Yu, Jiashuo and Zhang, Fan and Si, Chenyang and Jiang, Yuming and Zhang, Yuanhan and Wu, Tianxing and Jin, Qingyang and Chanpaisit, Nattapol and Wang, Yaohui and Chen, Xinyuan and Wang, Limin and Lin, Dahua and Qiao, Yu and Liu, Ziwei},
    title     = {VBench: Comprehensive Benchmark Suite for Video Generative Models},
    booktitle = {CVPR},
    month     = {June},
    year      = {2024},
    pages     = {21807-21818}
}

@inproceedings{wang2025koala,
  title={Koala-36m: A large-scale video dataset improving consistency between fine-grained conditions and video content},
  author={Wang, Qiuheng and Shi, Yukai and Ou, Jiarong and Chen, Rui and Lin, Ke and Wang, Jiahao and Jiang, Boyuan and Yang, Haotian and Zheng, Mingwu and Tao, Xin and others},
  booktitle={CVPR},
  pages={8428--8437},
  year={2025}
}

@article{ju2024miradata,
  title={Miradata: A large-scale video dataset with long durations and structured captions},
  author={Ju, Xuan and Gao, Yiming and Zhang, Zhaoyang and Yuan, Ziyang and Wang, Xintao and Zeng, Ailing and Xiong, Yu and Xu, Qiang and Shan, Ying},
  journal={NeurIPS},
  volume={37},
  pages={48955--48970},
  year={2024}
}

@article{wan2.1,
  title={Wan: Open and advanced large-scale video generative models},
  author={Wan, Team and Wang, Ang and Ai, Baole and Wen, Bin and Mao, Chaojie and Xie, Chen-Wei and Chen, Di and Yu, Feiwu and Zhao, Haiming and Yang, Jianxiao and others},
  journal={arXiv preprint arXiv:2503.20314},
  year={2025}
}

@article{seawead2025seaweed,
  title={Seaweed-7b: Cost-effective training of video generation foundation model},
  author={Seawead, Team and Yang, Ceyuan and Lin, Zhijie and Zhao, Yang and Lin, Shanchuan and Ma, Zhibei and Guo, Haoyuan and Chen, Hao and Qi, Lu and Wang, Sen and others},
  journal={arXiv preprint arXiv:2504.08685},
  year={2025}
}

@article{schuhmann2021laion,
  title={Laion-400m: Open dataset of clip-filtered 400 million image-text pairs},
  author={Schuhmann, Christoph and Vencu, Richard and Beaumont, Romain and Kaczmarczyk, Robert and Mullis, Clayton and Katta, Aarush and Coombes, Theo and Jitsev, Jenia and Komatsuzaki, Aran},
  journal={arXiv preprint arXiv:2111.02114},
  year={2021}
}

@inproceedings{raft,
  title={Raft: Recurrent all-pairs field transforms for optical flow},
  author={Teed, Zachary and Deng, Jia},
  booktitle={ECCV},
  pages={402--419},
  year={2020},
  organization={Springer}
}

@misc{pexels,
  title = {Pexels},
  year = {2025.10},
  howpublished = {\url{https://www.pexels.com/}},
  note = {accessed: 2025-11-01}
}

@inproceedings{
chefer2025videojam,
title={Video{JAM}: Joint Appearance-Motion Representations for Enhanced Motion Generation in Video Models},
author={Hila Chefer and Uriel Singer and Amit Zohar and Yuval Kirstain and Adam Polyak and Yaniv Taigman and Lior Wolf and Shelly Sheynin},
booktitle={ICML},
year={2025},
url={https://openreview.net/forum?id=yMJcHWcb2Z}
}

@article{hps-v2,
  title={Human preference score v2: A solid benchmark for evaluating human preferences of text-to-image synthesis},
  author={Wu, Xiaoshi and Hao, Yiming and Sun, Keqiang and Chen, Yixiong and Zhu, Feng and Zhao, Rui and Li, Hongsheng},
  journal={arXiv preprint arXiv:2306.09341},
  year={2023}
}

@article{xu2023imagereward,
  title={Imagereward: Learning and evaluating human preferences for text-to-image generation},
  author={Xu, Jiazheng and Liu, Xiao and Wu, Yuchen and Tong, Yuxuan and Li, Qinkai and Ding, Ming and Tang, Jie and Dong, Yuxiao},
  journal={NeurIPS},
  volume={36},
  pages={15903--15935},
  year={2023}
}

@article{pickscore,
  title={Pick-a-pic: An open dataset of user preferences for text-to-image generation},
  author={Kirstain, Yuval and Polyak, Adam and Singer, Uriel and Matiana, Shahbuland and Penna, Joe and Levy, Omer},
  journal={NeurIPS},
  volume={36},
  pages={36652--36663},
  year={2023}
}

@article{guo2023animatediff,
  title={AnimateDiff: Animate Your Personalized Text-to-Image Diffusion Models without Specific Tuning},
  author={Guo, Yuwei and Yang, Ceyuan and Rao, Anyi and Liang, Zhengyang and Wang, Yaohui and Qiao, Yu and Agrawala, Maneesh and Lin, Dahua and Dai, Bo},
  journal={ICLR},
  year={2024}
}

@inproceedings{unet,
  title={U-net: Convolutional networks for biomedical image segmentation},
  author={Ronneberger, Olaf and Fischer, Philipp and Brox, Thomas},
  booktitle={MICCAI},
  pages={234--241},
  year={2015},
  organization={Springer}
}

@inproceedings{diffusionDPO,
  title={Diffusion model alignment using direct preference optimization},
  author={Wallace, Bram and Dang, Meihua and Rafailov, Rafael and Zhou, Linqi and Lou, Aaron and Purushwalkam, Senthil and Ermon, Stefano and Xiong, Caiming and Joty, Shafiq and Naik, Nikhil},
  booktitle={CVPR},
  pages={8228--8238},
  year={2024}
}

@inproceedings{videoscore,
  title={Videoscore: Building automatic metrics to simulate fine-grained human feedback for video generation},
  author={He, Xuan and Jiang, Dongfu and Zhang, Ge and Ku, Max and Soni, Achint and Siu, Sherman and Chen, Haonan and Chandra, Abhranil and Jiang, Ziyan and Arulraj, Aaran and others},
  booktitle={EMNLP},
  pages={2105--2123},
  year={2024}
}

@inproceedings{vit,
  title={An Image is Worth 16x16 Words: Transformers for Image Recognition at Scale},
  author={Dosovitskiy, Alexey and Beyer, Lucas and Kolesnikov, Alexander and Weissenborn, Dirk and Zhai, Xiaohua and Unterthiner, Thomas and Dehghani, Mostafa and Minderer, Matthias and Heigold, Georg and Gelly, Sylvain and others},
  booktitle={ICLR},
  year={2020}
}

@inproceedings{sd3,
  title={Scaling rectified flow transformers for high-resolution image synthesis},
  author={Esser, Patrick and Kulal, Sumith and Blattmann, Andreas and Entezari, Rahim and M{\"u}ller, Jonas and Saini, Harry and Levi, Yam and Lorenz, Dominik and Sauer, Axel and Boesel, Frederic and others},
  booktitle={Forty-first international conference on machine learning},
  year={2024}
}

@inproceedings{dit,
  title={Scalable diffusion models with transformers},
  author={Peebles, William and Xie, Saining},
  booktitle={ICCV},
  pages={4195--4205},
  year={2023}
}

@article{visionreward,
  title={Visionreward: Fine-grained multi-dimensional human preference learning for image and video generation},
  author={Xu, Jiazheng and Huang, Yu and Cheng, Jiale and Yang, Yuanming and Xu, Jiajun and Wang, Yuan and Duan, Wenbo and Yang, Shen and Jin, Qunlin and Li, Shurun and others},
  journal={arXiv preprint arXiv:2412.21059},
  year={2024}
}

@article{attention,
  title={Attention is all you need},
  author={Vaswani, Ashish and Shazeer, Noam and Parmar, Niki and Uszkoreit, Jakob and Jones, Llion and Gomez, Aidan N and Kaiser, {\L}ukasz and Polosukhin, Illia},
  journal={NeurIPS},
  volume={30},
  year={2017}
}

@article{hu2022lora,
  title={Lora: Low-rank adaptation of large language models.},
  author={Hu, Edward J and Shen, Yelong and Wallis, Phillip and Allen-Zhu, Zeyuan and Li, Yuanzhi and Wang, Shean and Wang, Lu and Chen, Weizhu and others},
  journal={ICLR},
  volume={1},
  number={2},
  pages={3},
  year={2022}
}

@inproceedings{stablediffusion,
  title={High-resolution image synthesis with latent diffusion models},
  author={Rombach, Robin and Blattmann, Andreas and Lorenz, Dominik and Esser, Patrick and Ommer, Bj{\"o}rn},
  booktitle={CVPR},
  pages={10684--10695},
  year={2022}
}

@inproceedings{sdxl,
  title={Sdxl: Improving latent diffusion models for high-resolution image synthesis},
  author={Podell, Dustin and English, Zion and Lacey, Kyle and Blattmann, Andreas and Dockhorn, Tim and M{\"u}ller, Jonas and Penna, Joe and Rombach, Robin},
  booktitle={ICLR},
  year={2024}
}

@article{wang2025lavie,
  title={Lavie: High-quality video generation with cascaded latent diffusion models},
  author={Wang, Yaohui and Chen, Xinyuan and Ma, Xin and Zhou, Shangchen and Huang, Ziqi and Wang, Yi and Yang, Ceyuan and He, Yinan and Yu, Jiashuo and Yang, Peiqing and others},
  journal={IJCV},
  volume={133},
  number={5},
  pages={3059--3078},
  year={2025},
  publisher={Springer}
}

@article{yang2025ipo,
  title={IPO: Iterative preference optimization for text-to-video generation},
  author={Yang, Xiaomeng and Tan, Zhiyu and Li, Hao},
  journal={arXiv preprint arXiv:2502.02088},
  year={2025}
}

@inproceedings{song2020score,
  title={Score-Based Generative Modeling through Stochastic Differential Equations},
  author={Song, Yang and Sohl-Dickstein, Jascha and Kingma, Diederik P and Kumar, Abhishek and Ermon, Stefano and Poole, Ben},
  booktitle={International Conference on Learning Representations},
  year={2021}
}

@inproceedings{flowmatching,
  title={Flow Matching for Generative Modeling},
  author={Lipman, Yaron and Chen, Ricky TQ and Ben-Hamu, Heli and Nickel, Maximilian and Le, Matthew},
  booktitle={ICLR},
  year={2023}
}

@inproceedings{ddim,
  title={Denoising Diffusion Implicit Models},
  author={Song, Jiaming and Meng, Chenlin and Ermon, Stefano},
  booktitle={ICLR},
  year={2021}
}

@article{ddpm,
  title={Denoising diffusion probabilistic models},
  author={Ho, Jonathan and Jain, Ajay and Abbeel, Pieter},
  journal={NeurIPS},
  volume={33},
  pages={6840--6851},
  year={2020}
}

@inproceedings{hong2022cogvideo,
  title={CogVideo: Large-scale Pretraining for Text-to-Video Generation via Transformers},
  author={Hong, Wenyi and Ding, Ming and Zheng, Wendi and Liu, Xinghan and Tang, Jie},
  booktitle={ICLR},
  year={2023}
}

@inproceedings{vae,
  title={Auto-encoding variational bayes},
  author={Kingma, Diederik P and Welling, Max},
  booktitle={ICLR},
  year={2014}
}

@inproceedings{realesrgan,
  title={Real-esrgan: Training real-world blind super-resolution with pure synthetic data},
  author={Wang, Xintao and Xie, Liangbin and Dong, Chao and Shan, Ying},
  booktitle={ICCV},
  pages={1905--1914},
  year={2021}
}

@article{imagen,
  title={Photorealistic text-to-image diffusion models with deep language understanding},
  author={Saharia, Chitwan and Chan, William and Saxena, Saurabh and Li, Lala and Whang, Jay and Denton, Emily L and Ghasemipour, Kamyar and Gontijo Lopes, Raphael and Karagol Ayan, Burcu and Salimans, Tim and others},
  journal={NeurIPS},
  volume={35},
  pages={36479--36494},
  year={2022}
}

@article{dall-e2,
  title={Hierarchical text-conditional image generation with clip latents},
  author={Ramesh, Aditya and Dhariwal, Prafulla and Nichol, Alex and Chu, Casey and Chen, Mark},
  journal={arXiv preprint arXiv:2204.06125},
  volume={1},
  number={2},
  pages={3},
  year={2022}
}

@article{hunyuanvideo,
  title={Hunyuanvideo: A systematic framework for large video generative models},
  author={Kong, Weijie and Tian, Qi and Zhang, Zijian and Min, Rox and Dai, Zuozhuo and Zhou, Jin and Xiong, Jiangfeng and Li, Xin and Wu, Bo and Zhang, Jianwei and others},
  journal={arXiv preprint arXiv:2412.03603},
  year={2024}
}

@article{vdm,
  title={Video diffusion models},
  author={Ho, Jonathan and Salimans, Tim and Gritsenko, Alexey and Chan, William and Norouzi, Mohammad and Fleet, David J},
  journal={NeurIPS},
  volume={35},
  pages={8633--8646},
  year={2022}
}

@inproceedings{blattmann2023align,
  title={Align your latents: High-resolution video synthesis with latent diffusion models},
  author={Blattmann, Andreas and Rombach, Robin and Ling, Huan and Dockhorn, Tim and Kim, Seung Wook and Fidler, Sanja and Kreis, Karsten},
  booktitle={CVPR},
  pages={22563--22575},
  year={2023}
}

@article{ho2022imagenvideo,
  title={Imagen video: High definition video generation with diffusion models},
  author={Ho, Jonathan and Chan, William and Saharia, Chitwan and Whang, Jay and Gao, Ruiqi and Gritsenko, Alexey and Kingma, Diederik P and Poole, Ben and Norouzi, Mohammad and Fleet, David J and others},
  journal={arXiv preprint arXiv:2210.02303},
  year={2022}
}

@inproceedings{wu2023tune,
  title={Tune-a-video: One-shot tuning of image diffusion models for text-to-video generation},
  author={Wu, Jay Zhangjie and Ge, Yixiao and Wang, Xintao and Lei, Stan Weixian and Gu, Yuchao and Shi, Yufei and Hsu, Wynne and Shan, Ying and Qie, Xiaohu and Shou, Mike Zheng},
  booktitle={ICCV},
  pages={7623--7633},
  year={2023}
}

@inproceedings{khachatryan2023text2video,
  title={Text2video-zero: Text-to-image diffusion models are zero-shot video generators},
  author={Khachatryan, Levon and Movsisyan, Andranik and Tadevosyan, Vahram and Henschel, Roberto and Wang, Zhangyang and Navasardyan, Shant and Shi, Humphrey},
  booktitle={ICCV},
  pages={15954--15964},
  year={2023}
}

@article{zhou2022magicvideo,
  title={Magicvideo: Efficient video generation with latent diffusion models},
  author={Zhou, Daquan and Wang, Weimin and Yan, Hanshu and Lv, Weiwei and Zhu, Yizhe and Feng, Jiashi},
  journal={arXiv preprint arXiv:2211.11018},
  year={2022}
}

@inproceedings{ling2025vmbench,
  title={Vmbench: A benchmark for perception-aligned video motion generation},
  author={Ling, Xinran and Zhu, Chen and Wu, Meiqi and Li, Hangyu and Feng, Xiaokun and Yang, Cundian and Hao, Aiming and Zhu, Jiashu and Wu, Jiahong and Chu, Xiangxiang},
  booktitle={ICCV},
  pages={13087--13098},
  year={2025}
}

@inproceedings{nan2024openvid,
  title={OpenVid-1M: A Large-Scale High-Quality Dataset for Text-to-video Generation},
  author={Nan, Kepan and Xie, Rui and Zhou, Penghao and Fan, Tiehan and Yang, Zhenheng and Chen, Zhijie and Li, Xiang and Yang, Jian and Tai, Ying},
  booktitle={ICLR},
  year={2025}
}

@inproceedings{bain2021frozen,
  title={Frozen in time: A joint video and image encoder for end-to-end retrieval},
  author={Bain, Max and Nagrani, Arsha and Varol, G{\"u}l and Zisserman, Andrew},
  booktitle={ICCV},
  pages={1728--1738},
  year={2021}
}

@inproceedings{chen2024panda,
  title={Panda-70m: Captioning 70m videos with multiple cross-modality teachers},
  author={Chen, Tsai-Shien and Siarohin, Aliaksandr and Menapace, Willi and Deyneka, Ekaterina and Chao, Hsiang-wei and Jeon, Byung Eun and Fang, Yuwei and Lee, Hsin-Ying and Ren, Jian and Yang, Ming-Hsuan and others},
  booktitle={CVPR},
  pages={13320--13331},
  year={2024}
}

@inproceedings{chen2024internvl,
  title={Internvl: Scaling up vision foundation models and aligning for generic visual-linguistic tasks},
  author={Chen, Zhe and Wu, Jiannan and Wang, Wenhai and Su, Weijie and Chen, Guo and Xing, Sen and Zhong, Muyan and Zhang, Qinglong and Zhu, Xizhou and Lu, Lewei and others},
  booktitle={CVPR},
  pages={24185--24198},
  year={2024}
}

@article{wang2024cogvlm,
  title={Cogvlm: Visual expert for pretrained language models},
  author={Wang, Weihan and Lv, Qingsong and Yu, Wenmeng and Hong, Wenyi and Qi, Ji and Wang, Yan and Ji, Junhui and Yang, Zhuoyi and Zhao, Lei and XiXuan, Song and others},
  journal={NeurIPS},
  volume={37},
  pages={121475--121499},
  year={2024}
}

@inproceedings{loshchilovdecoupled,
  title={Decoupled Weight Decay Regularization},
  author={Loshchilov, Ilya and Hutter, Frank},
  booktitle={ICLR},
  year={2019}
}

@article{zheng2023secrets,
  title={Secrets of rlhf in large language models part i: Ppo},
  author={Zheng, Rui and Dou, Shihan and Gao, Songyang and Hua, Yuan and Shen, Wei and Wang, Binghai and Liu, Yan and Jin, Senjie and Liu, Qin and Zhou, Yuhao and others},
  journal={arXiv preprint arXiv:2307.04964},
  year={2023}
}

@inproceedings{liu2024grounding,
  title={Grounding dino: Marrying dino with grounded pre-training for open-set object detection},
  author={Liu, Shilong and Zeng, Zhaoyang and Ren, Tianhe and Li, Feng and Zhang, Hao and Yang, Jie and Jiang, Qing and Li, Chunyuan and Yang, Jianwei and Su, Hang and others},
  booktitle={ECCV},
  pages={38--55},
  year={2024},
  organization={Springer}
}

@inproceedings{kirillov2023segment,
  title={Segment anything},
  author={Kirillov, Alexander and Mintun, Eric and Ravi, Nikhila and Mao, Hanzi and Rolland, Chloe and Gustafson, Laura and Xiao, Tete and Whitehead, Spencer and Berg, Alexander C and Lo, Wan-Yen and others},
  booktitle={ICCV},
  pages={4015--4026},
  year={2023}
}

@inproceedings{ravisam,
  title={SAM 2: Segment Anything in Images and Videos},
  author={Ravi, Nikhila and Gabeur, Valentin and Hu, Yuan-Ting and Hu, Ronghang and Ryali, Chaitanya and Ma, Tengyu and Khedr, Haitham and R{\"a}dle, Roman and Rolland, Chloe and Gustafson, Laura and others},
  booktitle={ICLR},
  year={2025}
}
